\pdfoutput=1
\documentclass[11pt]{article}

\usepackage[]{ACL}

\usepackage{times}
\usepackage{latexsym}
\usepackage{amsmath}

\usepackage{color}

\newcommand{\Blue}[1]{\textcolor[rgb]{0.00,0.00,1.00}{#1}}

\usepackage{cleveref}
\crefformat{section}{\S#2#1#3}
\crefformat{subsection}{\S#2#1#3}
\crefformat{subsubsection}{\S#2#1#3}
\crefrangeformat{section}{\S#3#1#4 to~\S#5#2#6}
\crefmultiformat{section}{\S#2#1#3}{ and~\S#2#1#3}{, #2#1#3}{ and~#2#1#3}
\Crefformat{figure}{#2Fig.~#1#3}
\Crefmultiformat{figure}{Figs.~#2#1#3}{ and~#2#1#3}{, #2#1#3}{ and~#2#1#3}
\Crefformat{table}{#2Tab.~#1#3}
\Crefmultiformat{table}{Tabs.~#2#1#3}{ and~#2#1#3}{, #2#1#3}{ and~#2#1#3}
\Crefformat{appendix}{#2Appx.~\S#1#3}
\usepackage[T1]{fontenc}
\usepackage[utf8]{inputenc}

\usepackage{microtype}

\usepackage{times}
\usepackage{latexsym}

\usepackage{array}
\usepackage{booktabs}
\usepackage{multirow}
\usepackage{amssymb}
\usepackage{graphicx}
\usepackage{enumerate}
\usepackage{diagbox}
\usepackage{microtype}
\usepackage{algorithm}
\usepackage{algpseudocode}
\newcommand{\tabincell}[2]{\begin{tabular}{@{}#1@{}}#2\end{tabular}} 
\usepackage{color}

\usepackage{inconsolata}

\title{Getting Sick After Seeing a Doctor? Diagnosing and Mitigating \\Knowledge Conflicts in Event Temporal Reasoning}

\author{Tianqing Fang$^{1}$, Zhaowei Wang$^{1}$, Wenxuan Zhou$^{2}$, Hongming Zhang$^{3}$,\\
\textbf{Yangqiu Song$^{1}$, Muhao Chen$^{2}$}\\
$^{1}$Hong Kong University of Science and Technology~~~$^{2}$University of Southern California\\
$^{3}$Tencent AI Lab, Seattle~~~$^{4}$University of California, Davis \\
 \texttt{\{tfangaa, zwanggy, yqsong\}@cse.ust.hk}, \texttt{zhouwenx@usc.edu}\\ 
 \texttt{hongmzhang@global.tencent.com}, \texttt{muhchen@ucdavis.edu}
\\
}

\begin{document}
\maketitle
\begin{abstract}
Event temporal reasoning aims at identifying the temporal relations between two or more events from narratives.
However, \textit{knowledge conflicts} arise when there is a mismatch between the actual temporal relations of events in the context and the prior knowledge or biases learned by the model.
In this paper, we propose to detect knowledge-conflict examples in event temporal reasoning using bias indicators,
which include \textit{event relation prior} bias, \textit{tense} bias, \textit{narrative} bias, and \textit{dependency} bias.
We define conflict examples as those where event relations are opposite to biased or prior relations.
To mitigate 
% such 
event-related knowledge conflicts, we introduce a Counterfactual Data Augmentation (CDA) based method that can be applied to both Pre-trained Language Models (PLMs) and Large Language Models (LLMs) either as additional training data or demonstrations for In-Context Learning. 
Experiments suggest both PLMs and LLMs suffer from knowledge conflicts in event temporal reasoning, and CDA has the potential for reducing \textit{hallucination} and improving model performance\footnote{Code and data are available at \url{https://github.com/tqfang/event-temporal-knowledge-conflict}}.
% Experiments suggest the importance of mitigating knowledge conflicts in event temporal reasoning tasks for reducing \textit{hallucination} and highlight the potential of counterfactual data augmentation for improving model performance. 

\end{abstract}

\section{Introduction}

%Event-level temporal reasoning aims at identifying
An important goal of event understanding is to identify the temporal relations (\textsc{TempRels}) among events described in natural language text \cite{chambers2007classifying}.
This task aligns with  human's cognitive ability~\cite{zacks2001event, zacks2007event}, which often involves routinely reasoning about 
%the temporal relations between events in order to make sense of the world around them, plan actions, and make decisions
how events happening around us are temporally sequenced, planned, and lead to consequences and decisions~\cite{schank1977scripts}. 
From the intelligent system perspective, it also
benefits many NLP applications for narrative understanding~\cite{li2018constructing, DBLP:conf/aaai/CaiDCDL22}, schema induction~\cite{li-etal-2021-future}, and question answering~\cite{DBLP:journals/ijcv/ZhuXYH17, stricker-2021-question-answering}.

\begin{figure}[t]
    \centering
    \includegraphics[width=0.95\linewidth]{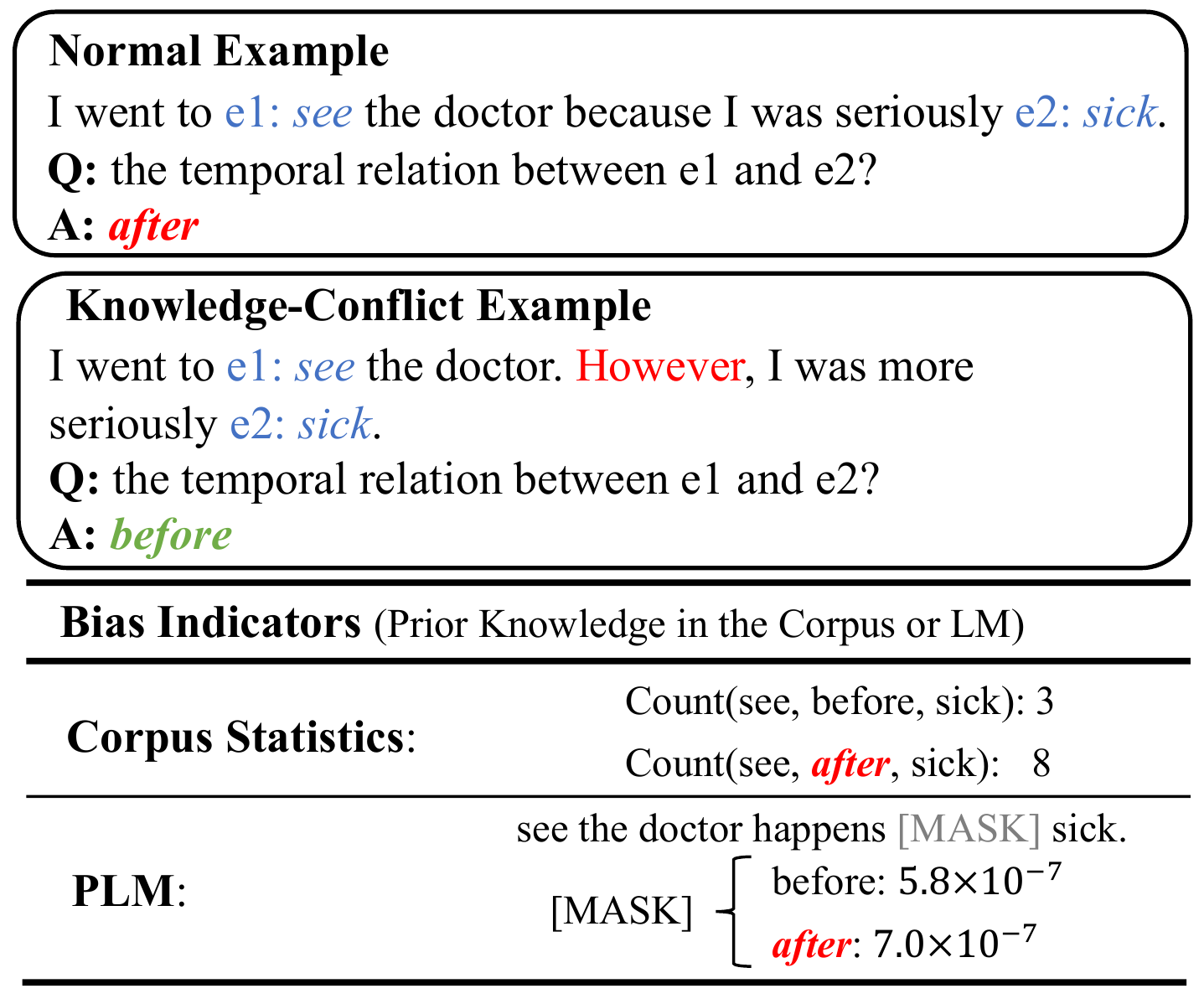}
    \vspace{-0.5em}
    \caption{
    An example of a knowledge-conflict instance. The actual \textsc{TempRel} in the context differs from the biased or prior \textsc{TempRel} in the corpus and the language model, leading to the emergence of \textit{knowledge conflicts}.
    }
    % \vspace{-1.5em}
    % Example of \textit{knowledge conflicts} in event temporal reasoning. 
    % The priors based on context-free corpus statistics or knowledge in language models may hinder the \textsc{TempRel} inference in context-sensitive scenarios.}
    \label{fig:intro}
\end{figure}

%In event temporal reasoning, context is the most important source that a model should refer to.
% If we look closely at what is presented to a temporal reasoning model, we may easily divide the input into two parts, including the event mentions themselves as well as the context.
In event temporal reasoning, the input includes two parts, the event mentions and the context.
% \muhao{this needs to point to an example and show what are event mentions, and what is the context; otherwise it could be confusing to readers.}
%It is widely recognized that
The \textsc{TempRel} a model seeks to infer should be based on the context, rather than only %from 
revealed by the event mentions themselves.
For example, in \Cref{fig:intro}, without a context, the event mention \textit{see} (the doctor) and \textit{sick} have certain temporal prior where \textit{see the doctor} %tend to statistically appear
statistically happen more often after \textit{sick}, either by corpus statistics or probing a masked PLM.
However, under the context of ``{I went to see the doctor, \textit{However}, I was more seriously sick},'' we can infer that \textit{see} happens before \textit{sick} instead of after due to the presence of the connective \textit{However}.
This is known as the phenomenon of \textit{knowledge conflicts}~\cite{longpre-2021-entity-knowledge-conflict}, where the contextual information contradicts the knowledge memorized by the language model.
Hence, the essential requirement for accountable temporal reasoning is
\textit{context-faithfulness}~\cite{DBLP:conf/eacl/WangZDGRC23, zhou2023context}, where models are expected to perform reasoning based on the context instead of \textit{guessing} using only the prior knowledge about the events encoded in their parameters.

However, most current language models, including both Pre-trained Language Models (PLMs) and Large Language Models (LLMs)\footnote{PLMs, or smaller models, are used in a pre-train and fine-tune paradigm, while LLMs, larger and more powerful models with over 10B parameters, are commonly employed through in-context learning~\cite{DBLP:journals/corr/abs-2303-09136}.}, 
rely on shortcuts from the mentions without being faithful to the context~\cite{DBLP:conf/emnlp/XuWLDC22, DBLP:conf/fat/BenderGMS21} to varying degrees, leading to \textit{hallucination}.
This issue is particularly severe in contexts where event or entity mentions have a different relation prior than what is presented in the context.
% To evaluate the extent that models suffer from such knowledge conflicts, we systematically define multiple kinds of knowledge conflicts %in event-level temporal reasoning. 
% related to events.
Though entity-related knowledge conflicts \cite{longpre-2021-entity-knowledge-conflict,DBLP:conf/naacl/WangCZCLLYLH22,li2022large} have recently attracted much attention, questions about event-related knowledge conflicts remained intact.

First, it is necessary to understand the conflicts regarding \textit{relations} of events, which is more complicated than that of a single event. 
Second, the substitution-based paradigm defined in entity knowledge conflicts or spurious correlation detection~\cite{longpre-2021-entity-knowledge-conflict} cannot be directly applied to events.
Entity mentions can often be replaced randomly with other entities with the same typing to study the faithfulness towards the context other than the entity mention, which remains unchanged after the replacement.
For example, in open-domain QA,
% \muhao{is this a specific Open-domain QA dataset that requests a citation?}, 
% there is a question ``What country did Germany invaded?'' based on the context ``Germany invaded \textit{Poland} in WWII.''
% To check whether models faithfully rely on the context instead of hallucinating, \textit{Poland} in the context can be changed to a random country name to see if the model can still output the correct country instead of \textit{Poland} as they have learned from either pre-training or fine-tuning.
a possible question can be ``Who is the CEO of Twitter?'' based on the context ``Yaccarino succeeded Elon Musk as the CEO of Twitter''.
To check whether models faithfully rely on the context instead of hallucinating, \textit{Yaccarino} in the context can be changed to a random name to see if the model can still output the ``correct'' CEO instead of \textit{Yaccarino} as they have learned in pre-training.
% However, events are usually denoted by predicates in the context, and directly substituting the predicate (e.g., from \textit{see} in \Cref{fig:intro} to another random verb \textit{play}) will alter the semantic meaning of the whole context, including both predicate and its dependency with the arguments, making it infeasible to analyze the \textit{faithfulness} towards the \textit{original} context as a whole.
However, events are usually denoted by predicates in the context~\cite{DBLP:conf/semco/BethardMK07}, and directly substituting the predicate (e.g., from \textit{see} in \Cref{fig:intro} to another random verb such as \textit{play}) will alter the semantic meaning of the whole context, including both the predicate and its dependency with the arguments, making it infeasible to analyze the \textit{faithfulness} towards the \textit{original} context.
% cannot be trivially substituted without altering their arguments, making it hard to evaluate the faithfulness to the \textit{original} context as a whole.
% Take the case in \Cref{fig:intro} as an example, 
%predicates
% and cannot be easily substituted without corrupting the semantics.
% \muhao{what's this? +citations}.
% However, events are more dominant components of the context and cannot be trivially substituted. 
Thus, instead of resorting to a substitution, in this paper, we study the effect of knowledge conflicts in event temporal reasoning by 
\textit{selecting} conflict examples from the original dataset based on corpus statistics, and evaluate models on the conflict subsets.

We outline the contributions of this paper as follows.
% In this paper, we present a comprehensive framework for measuring knowledge conflicts in event temporal reasoning, 
First, we define four types of bias that can lead to knowledge conflicts, including \textit{event-relation} bias, \textit{narrative} bias, \textit{tense} bias, and \textit{dependency} bias.
The data instances where the actual \textsc{TempRel} contradicts with the 
prior \textsc{TempRel} 
% biased or prior \textsc{TempRel} 
are referred to as knowledge-conflict instances (\Cref{sec:knowledge_conflict_detection}), as they conflict with the prior knowledge provided to language models.
% Different from substitution-based methods~\cite{longpre-2021-entity-knowledge-conflict}, we directly \textit{select} a subset from the original evaluation set based on bias statistics that represents the degree of conflicts (\Cref{sec:knowledge_conflict_detection}).
Second, to mitigate the effect of knowledge conflicts, we propose a Counterfactual Data Augmentation (CDA) technique that explicitly generates contexts with knowledge-conflict elements, thereby reducing the overall bias in the data distribution.
CDA can be applied to both fine-tuned PLMs and LLMs with (test-time) in-context learning (\Cref{sec:knowledge_conflict_mitigation}). 
Third, 
we study the effect of various kinds of knowledge conflicts and our proposed bias mitigation method on two popular event temporal reasoning benchmarks,
% we adopt our methods on two popular event temporal reasoning benchmarks, 
TORQUE~\cite{DBLP:conf/emnlp/NingWHPGR20torque} and MATRES~\cite{DBLP:conf/acl/RothWN18matres}. 
We show that models suffer from performance drop on knowledge-conflict subsets, and our bias-aware data augmentation method outperforms baselines by a remarkable margin on both bias mitigation and overall performance (\Cref{sec:experiments}).

% It happens as a result of over-reliance on memorized information that contains \textit{reporting bias}~\cite{DBLP:conf/cikm/GordonD13}.

\section{Related Works}

\paragraph{Event Temporal Reasoning. } 
Event temporal reasoning aims at identifying the temporal relations (\textsc{TempRel}) of events in narratives.
There are two common ways of formulating this problem.
The first formulation is the \textsc{TempRel} extraction task, which involves determining the \textsc{TempRel} between two annotated event triggers from a pre-defined relation set~\cite{DBLP:conf/semco/BethardMK07, DBLP:conf/semeval/BethardSPP17, DBLP:conf/acl/RothWN18matres, DBLP:conf/sigdial/NaikBR19}.
Meanwhile, another formulation is a reading comprehension task, which involves determining more complicated \textsc{TempRels} expressed in natural language questions~\cite{DBLP:conf/emnlp/NingWHPGR20torque, DBLP:conf/emnlp/HanHSBNRP21}. 
To conduct event temporal reasoning, %a pile of works are done to
literature has leveraged various approaches, 
including graph neural networks
% on event dependency relation sub-graphs
~\cite{DBLP:conf/naacl/ZhangNH22, DBLP:conf/coling/ZhouDTWD22}, 
rhetorical discourse features and temporal arguments from semantic role labels~\cite{mathur-etal-2021-timers}, 
distant supervision
% using supervision signals provided by per-trained semantic role labeling techniques
~\cite{zhou-etal-2021-temporal, zhao-etal-2021-effective},
and event relation joint learning~\cite{DBLP:conf/emnlp/WangCZR20, DBLP:conf/acl/0003DZ0WFSWS23, DBLP:conf/emnlp/WangZFSWS22}.
% In addition, \citet{DBLP:conf/eacl/WangZDGRC23} study the effect of counterfactual inference
% as well as 
% Dirichlet parameterization to improve uncertainty calibration of the model.
LLMs such as GPT3~\cite{DBLP:conf/nips/BrownMRSKDNSSAA20} and ChatGPT are also leveraged for event temporal reasoning~\cite{DBLP:journals/corr/abs-2304-14827} with carefully designed prompts and In-Context Learning.
% However, these methods only focus on how to leverage existing features beneficial to event temporal reasoning without considering the the impact of dataset bias and knowledge conflicts of the models. 
Our work differs from previous studies in that we study the knowledge conflicts in event temporal reasoning and how to mitigate them.

\paragraph{Knowledge Conflict in Language Models.}
Knowledge conflicts have been widely studied for entity-centric NLU tasks~\cite{schuster-etal-2021-get, DBLP:journals/tvcg/WangHJFQ24}.
For example, \citet{longpre-2021-entity-knowledge-conflict} studied the knowledge conflict in open-domain question answering using entity substitution.
\citet{li2022large} also adopted this strategy to study the enhancement of a PLM's robustness against context noise with a knowledge-aware working memory.
% For example, it substitutes an answer entity from the context with another entity of the same type, so as to check whether the language model gives faithful predictions based on the context or it generates hallucinations. 
\citet{schuster-etal-2021-get} proposed a dataset using the user edits in Wikipedia as subtle changes for the context and study the effect of such changes on language models.
\citet{DBLP:conf/emnlp/XuWLDC22} systematically formulate six types of biases in entity typing to study spurious correlations. 
Certain types of biases, such as Mention-Context and Named Entity bias, can reflect knowledge conflicts in entities. 
% In these cases, the model may hallucinate based solely on the entity span that contains rich semantics, ignoring the surrounding context. 
%\citet{DBLP:journals/corr/abs-2211-14358} studied event-centric bias towards different genders in fairy tales.
% \citet{zhou2023context} use opinion-based prompting and counterfactual demonstration to enhance the context-faithfulness of test-time-only LLMs against knowledge conflicts.
Typical mitigation methods of knowledge conflicts include causal analysis and coutnerfactual analysis~\cite{DBLP:conf/emnlp/WangMWZC23, DBLP:conf/naacl/WangCZCLLYLH22}, and test-time-only LLM prompting methods using counterfactual demonstration and opinion-based prompting~\cite{zhou2023context}.
However, the knowledge conflicts of event-event \textsc{TempRels} are under-explored. 
\citet{DBLP:journals/corr/abs-2212-10467} proposed a dataset studying the differential effects of \textsc{TempRel} reasoning given additional contexts, while their focus is on annotating additional out-of-distribution data instead of exploring existing knowledge conflicts within the dataset.
Our work systematically defines and detects knowledge conflicts in event temporal reasoning and proposes a data-augmentation-based method to mitigate those conflicts based on the detected bias.

\begin{table*}[t]
\centering
\small
\renewcommand\arraystretch{1.1}
\newcolumntype{C}[1]{>{\centering\arraybackslash}p{#1}}
\begin{tabular}{l|p{29em}c}
\toprule
  Type & \centering{Context \& Label} & Bias Scores \\
 \midrule
 % \tabincell{l}{Relation Prior \\ (relation)} & \tabincell{p{30em}}{(TORQUE) Ecuador's Rafael Correa has \Blue{$e_1$: won} Sunday's presidential \Blue{$e_2$: run-off} with 57.04 percent of the \Blue{$e_3$: votes} against his rival Alvaro Noboa with 42.96 percent, after 97.29 percent of \Blue{$e_4$: votes} were \Blue{$e_5$: counted}, Ecuador's Supreme Electoral Tribunal \Blue{$e_6$: said} on Tuesday. \\ \textbf{Question}: What happened after the \Blue{$e_3$: votes}? \\ \textbf{True label}: \Blue{$e_1$: won}, $e_5$: counted, $e_6$: said \\ \textbf{Biased Prediction}: $e_5$: counted, $e_6$: said  } & \tabincell{l}{$b(\textit{vote}, \textit{win}, \textit{before}) = 1.0$ \\ (\textit{frequency} $= 7$)} \\
 \tabincell{l}{Relation Prior \\ (relation)} & \tabincell{p{30em}}{(TORQUE) Chidambaram \Blue{$e_1$:drew} up the previous United Front government's Indian budget for 1997-98 which is to be \Blue{$e_2$: approved} by parliament this week . Gujral has \Blue{$e_3$: adopted} the same budget.\\ \textbf{Question:} What will happen after \Blue{$e_1$: drew}? \\ \textbf{True label:} \Blue{$e_2$: approve}. \textbf{Biased Prediction:} $e_3$: adopted } &  \tabincell{l}{$b(\textit{draw}, \textit{adopt}, \textit{before}) = 1.0$ \\ $b(\textit{draw}, \textit{approve}, \textit{before}) = 0$} \\
 \cline{2-3}
 % \tabincell{l}{Relation Prior\\(warm-up)} & \tabincell{p{30em}}{\textbf{Question}: What event has already finished? \\ \textbf{True label}: $e_1$, $e_2$, $e_3$, $e_4$, \Blue{$e_5$: counted}, $e_6$ \\ \textbf{Biased Prediction}: $e_1$, $e_2$, $e_3$, $e_4$, $e_6$} & \tabincell{l}{$b(\textit{count}, \textit{happened}) = 0.4$ \\ $b(\textit{count}, \textit{future}) = 0.6$ \\ (\textit{frequency} $= 5$)} \\
 \tabincell{l}{Relation Prior\\(warm-up)} &  \tabincell{p{30em}}{\textbf{Question:} What will happen in the future? \\ \textbf{True label: } \Blue{$e_2$: approve}. \textbf{Biased Prediction: }{$e_3$: adopt} } & \tabincell{l}{ $b(\textit{approve}, \textit{happened})=0.9$ \\ $b(\textit{approve}, \textit{future}) = 0.05$} \\
 \hline
\tabincell{l}{Tense \\ (relation)} &\tabincell{p{30em}}{(MATRES) Albright \Blue{$e_1$: told (\texttt{VBD})} ambassadors 
% and charges d'affaires of 30 African countries 
of 30 African countries 
in Washington, who came to the State Department to \Blue{$e_2$: offer (\texttt{VB})} condolences. \\ \textbf{True label}: $e_1$ happens \textit{after} $e_2$; \textbf{Biased Prediction:} \textit{before}  } & \tabincell{l}{$b(\texttt{VBD}, \texttt{VB}, \textit{before}) = 0.70$ \\ $b(\texttt{VBD}, \texttt{VB}, \textit{after})\ \  = 0.27$\\$b(\texttt{VBD}, \texttt{VB}, \textit{equal})\ = 0.03$} \\
\cline{2-3}
\tabincell{l}{Tense\\(warm-up)} & \tabincell{p{30em}}{(TORQUE) That's what will \Blue{$e_1$: keep} computer makers \Blue{$e_2$: coming (\texttt{VBG})} in spite of the \Blue{$e_3$: irritation} of \Blue{$e_4$: bugs}. \\ \textbf{Question:} What will happen in the future? \\ \textbf{True Label:} $e_1$, \Blue{$e_2$: coming}; \textbf{Biased Prediction}: $e_1$ } & \tabincell{l}{$b(\texttt{VBG}, \textit{happened}) = 0.42$ \\ $b(\texttt{VBG}, \textit{future}) = 0.13$\\ $b(\texttt{VBG}, \textit{happening}) = 0.45$}\\
\hline
Narrative & \tabincell{p{30em}}{(MATRES) Now events are \Blue{$e_1$: doing} the work for Schumer. Slepian's death was among the first topics \Blue{$e_2$: raised} in Saturday night's debate between the two men, ...
% and it was instantly followed by a question to D'Amato on whether he opposed first-trimester abortions for adult women. 
;\ \textbf{True label}: $e_1$ happens \textit{after} $e_2$; \textbf{Biased Prediction:} \textit{before} } & \tabincell{l}{$b(p_1 < p_2, \textit{before}) = 0.59$ \\ $b(p_1 < p_2, \textit{after})\ \  = 0.37$\\$b(p_1 < p_2, \textit{equal})\ = 0.04$} \\
\hline
Dependency & \tabincell{p{30em}}{(MATRES) Castro \Blue{$e_1$: said} Gonzalez would \Blue{$e_2$: travel} with his current wife and their son
% , a top government official, doctors, nurses, psychiatrists, ...
% Elian's Cuban kindergarten teacher, classmates and his old school desk. 
(Dependency: \textit{says} $\rightarrow$ \texttt{ccomp} $\rightarrow$ \textit{travel}) \\ \textbf{True label}: $e_1$ happens \textit{before} $e_2$; \textbf{Biased Prediction:} \textit{after} } & \tabincell{l}{$b(\texttt{ccomp}, \textit{before}) = 0.66$ \\ $b(\texttt{ccomp}, \textit{after})\ \  = 0.32$ \\ $b(\texttt{ccomp}, \textit{equal})\ = 0.02$}  \\
\bottomrule
\end{tabular}
% \vspace{-0.1in}
\caption{Examples of different forms of knowledge conflicts. } \label{tab:bias_examples}
% \vspace{-1.5em}
\end{table*}
\section{Event Knowledge Conflict}\label{sec:knowledge_conflict_detection}

We introduce the problem definition (\Cref{sec:promblem_definition}) and
formally define four types of bias and how to select 
knowledge-conflict data % to study event-related knowledge conflicts 
(\Cref{sec:bias_diagnoses}).
% We show our defined bias is a versatile indicator for creating knowledge conflicts, allowing event relations from various sources and levels of conflicts.
We then introduce our proposed Counterfactual Data Augmentation for mitigating knowledge conflict (\Cref{sec:knowledge_conflict_mitigation}).

\subsection{Problem Definition}\label{sec:promblem_definition}
% In event temporal reasoning, the primary objective is to determine the \textsc{TempRel} between two or more events, 
% which previous studies~\cite{DBLP:conf/acl/RothWN18matres, DBLP:conf/sigdial/NaikBR19} typically classify as \textit{before}, \textit{after}, \textit{equal} (indicating two events occurring simultaneously), and \textit{vague}.
In event temporal reasoning, the primary objective is to determine the \textsc{TempRel} between two or more events.
Without the loss of generality, our study is based on pairwise event relations: the relation $r$ of an event pair $(e_1, e_2)$ based on the context $c$.
More complex cases can be easily addressed by breaking down the relations involving multiple events into pairwise relations.
For example, in TORQUE, a question might be formulated as ``what happens before $e$?'', where $e$ can be a target event trigger. 
Then, the answers $e_{a1}, e_{a2}, \cdots$ can form multiple event pairs ($e$, before, $e_{ai}$), where $i=1, 2, \cdots$, etc.
The case where evaluating the temporal status of a single event (\textit{happened}, \textit{happening}, \textit{will happen}, etc.) can also be easily adapted in this framework by replacing the features of event pairs to a single event.
% Adaptations to different datasets will be introduced in \Cref{sec:exp_conflict_diagnoses}.

% For example, in normal corpora, it's often the case when the event \textit{seeing a doctor} happens after \textit{getting sick}, while  \textit{getting sick} can definitely happens after \textit{seeing a doctor} as a result of contagion from the doctor.
% Different from substitution-based methods in entity-related knowledge conflicts, events are usually predicates in the context and cannot be trivially substituted without changing the context and the event arguments.
% As a result, we create an automated framework to detect event relation bias in datasets by first defining four types of bias statistics, 
To study event-related knowledge conflict, 
we create an automated framework to use corpus co-occurrence statistics to select conflict subsets.
Similar to the co-occurrence statistics in \textit{reporting bias}~\cite{DBLP:conf/cikm/GordonD13}, to obtain knowledge-conflict data, we first define bias, as the \textit{opposite} side of the conflict. 
We identify four types of bias in event temporal reasoning and defined corresponding bias statistics. We then selected a subset of the original dataset where feature-relation pairs were rare (i.e., \textit{knowledge-conflict}) based on the bias scores.
As the (reporting) bias in the training corpus is usually learned and amplified by the language models~\cite{DBLP:conf/coling/ShwartzC20}, 
% our selected subsets which stand as the opposite side of the bias are thus conflict with the knowledge encoded in the language models.
our selected subsets, which represent the opposite side of the bias, conflict with the knowledge encoded in the language models.
% We use various bias detection methods to select a subset of the evaluation set that counters the biased distribution in the training set as a subset to reflect conflicts of knowledge.

\subsection{Knowledge Conflict Diagnosis}\label{sec:bias_diagnoses}

% As we aim to identify the bias in event temporal reasoning about certain relations, 
We first define a bias score $b(P_1, P_2, r)$ with regard to certain patterns ($P_1$ and $P_2$) against a specific relation $r \in R$, where $R$ is a subset of all relations defined in a certain dataset. 
Patterns $P_i$ can be the event lemmas themselves, tense, dependency patterns, and narrative orders of either event. 
Sometimes $(P_1, P_2)$ is represented by one feature only, for example, the dependency relation and narrative orders between two events.
Denote $c(P_1, P_2, r)$ as the number of occurrences of ($P_1, P_2$) under relation $r$ in a corpus, and the bias score is defined as:
% \vspace{-0.5em}
\begin{equation}
    \small
    b(P_1, P_2, r) = \frac{c(P_1, P_2, r)}{\sum_{r'\in R} c(P_1, P_2, r')}
\end{equation}

For example, in tense bias, the bias score of the tense pattern (\texttt{VBD}, \texttt{VBZ}) (\textit{past tense} and \textit{third person singular present tense}) when only considering two relations $R=\{\text{\textit{before, after}}\}$ is defined as:
% \vspace{-0.5em}
\begin{align}
\small
\begin{aligned}
b(\texttt{VBD}, &\texttt{VBZ}, \textit{before}) \text{=} \frac{c(\texttt{VBD}, \texttt{VBZ}, \textit{before})}{c(\texttt{VBD}, \texttt{VBZ}, \textit{before}) \text{+} c(\texttt{VBD}, \texttt{VBZ}, \textit{after})}    
\end{aligned}
\end{align}

\paragraph{Knowledge Conflict Detection.}
% With bias scores, we can then select knowledge-conflict instances whose patterns do \textit{not} follow the majority distribution in the dataset.
In a set of relations, those with higher bias scores indicate higher degrees of bias towards certain relations, and others with lower bias scores indicate higher degrees of knowledge conflict.
We select instances whose patterns do \textit{not} follow the majority distribution in the training set as knowledge-conflict instances.
% A new pattern pair $(p_1, p_2)$ in the test set is considered knowledge-conflict if its actual relation is not the frequent relation according to the bias score $r \neq \text{argmax}_{r'\in R}\ \{b(p_1, p_2, r')\}$.
A new instance in the test set with a pattern-relation pair $(P_1, P_2, r)$ is considered knowledge conflict if the bias score is less than the context-free frequency of relations $b(P_1, P_2, r) < \frac{c(r)}{\sum_{r'\in R}c(r')}$.
Moreover, to ensure a significant degree of conflicts, we set a threshold $\mathcal{T}_r$ such that $b(P_1, P_2, r) < \mathcal{T}_r < \frac{c(r)}{\sum_{r'\in R}c(r')}$, to ensure that the conflict is large enough\footnote{Hyperparameter analysis on $\mathcal{T}_r$ is presented in \Cref{sec:appendix_bias_param}.}.
For example, a test instance where the event with a past tense happens \textit{after} the event with a present tense may be selected as a knowledge-conflict instance, as the context makes the actual \textsc{TempRel} different from the biased relation \textit{before}.

% We manually set a threshold to determine whether $(p_1, p_2, r)$ is biased or not. Generally, a pattern pair $(p_1, p_2)$ is considered bias if $\exists r \in R$ such that $b(p_1, p_2, r) > t_r \geq \frac{c(r)}{\sum_{r'\in R}c(r')}$. $t_r$ can be manually set to control the degree of bias. 
% We select the knowledge-conflict instances such that the selected 
% For example $b(\texttt{VBD}, \texttt{VBZ}, \textit{before})=0.7$

Next, we introduce the definitions of different forms of bias in detail. 
% Data instances that counteract the biased distribution are selected as corresponding knowledge-conflict subsets.

\paragraph{Relation Prior Bias.}
Bias toward certain \textsc{TempRels} exists because there are natural \textit{selectional preference}~\cite{DBLP:journals/ai/Wilks75} between the specific events.
For example, in the TORQUE dataset, \textit{arresting} %tends to 
dominantly happen after \textit{killing}, and \textit{voting} more often happens before \textit{winning}. 
These findings suggest that the occurrence of certain events may be more likely to follow or precede other events, which can 
however, lead to bias when the context describes the \textsc{TempRel} differently from the most frequent cases.
Our definition of the bias scoring function is based on the frequency of the co-occurrence of event $e_1$ and $e_2$ under relation $r$:
% \vspace{-0.5em}
\begin{equation}
\small
b(e_1, e_2, r) = \frac{c(e_1, e_2, r)}{\sum_{r'\in R} c(e_1, e_2, r')}    
\end{equation}

\paragraph{Narrative Bias.}
Narrative bias in event temporal reasoning is the tendency for the model to interpret the %order in which events are presented in a narrative as reflecting the actual temporal order of those events.
chronological order of the events to be the same as their narrative order.
However, these two orders, though more often accord with each other, do not always necessarily follow the same~\cite{zwaan1995dimensions}.
%\muhao{better + a citation}. %narrative order can lead to a spurious correlation that attributes to conflict predictions.
% However, the model spuriously pick up the narrative order to infer temporally.
In this sense, we only study \textit{before}, \textit{after}, and \textit{equal} relations for narrative bias.
Denote $p = P(e, c)$ as the position of event $e$ in context $c$, where the earlier position of $e$ indicates that this event is described earlier in the narrative.
The bias scoring function is defined as follows for the case where the positions of the two events follow the order of $p_1<p_2$:

% \vspace{-2em}
\begin{equation}
\small
b(p_1<p_2, \textit{before})=\frac{c(p_1<p_2, \textit{before})}{\sum_{r'\in R} c(p_1<p_2, r')}
\end{equation}

We select the event pairs where $p_1 < p_2$ while the actual relation is ($e_1$, \textit{after/equal}, $e_2$) or $p_1 > p_2$ while the actual relation is ($e_1$, before/equal, $e_2$) as the knowledge-conflict examples.

\paragraph{Tense Bias.}
Tense bias is the tendency to rely on the grammatical tense of verbs as evidence for the temporal order of events. 
For example, past tense is typically used to describe events that occurred \textit{before} the present moment, while present tense is typically used for events that are happening now or in the future. However, this grammatical convention does not always correspond to the actual temporal order of events. 
Denote $t_1$ and $t_2$ as the tense (POS-tags parsed by Spacy
\footnote{https://spacy.io/} 
as more fine-grained tense information) of event $e_1$ and $e_2$ under context $c$, then the bias score is defined as:
% \vspace{-0.5em}

\begin{equation}
    \small
    b(t_1, t_2, r) = \frac{c(t_1, t_2, r)}{\sum_{r'\in R} c(t_1, t_2, r')}
\end{equation}

\paragraph{Dependency Bias}

Dependency bias is the tendency to rely on syntactic dependency patterns in language as evidence for the temporal order of events.
For example, if two events $e_1$ and $e_2$ are directly connected in the dependency tree, 
the dependency pattern $(e_1, \texttt{dobj}, e_2)$ (where $e_1$ is the subject of the sentence, $e_2$ is the direct object, and \texttt{dobj} is the dependency between them) often indicates that $e_1$ is the entity performing an action on $e_2$. 
This pattern may suggest that $e_1$ must occur before $e_2$ in time, but this is not always the case.
Denote $d$ as the dependency relation between $e_1$ and $e_2$ in context $c$ ($d$ is null if $e_1$ and $e_2$ are not directly linked in their dependency tree).
% \vspace{-0.5em}

\begin{equation}
    \small
    b(d, r) = \frac{c(d, r)}{\sum_{r'\in R} c(d, r')}
\end{equation}

We summarize the core features of each defined bias associated with examples in \Cref{tab:bias_examples}.
Our focus is particularly on two datasets, namely TORQUE and MATRES, which will be presented in \Cref{sec:dataset_intro}. 
Prior to that, we introduce our proposed conflict-mitigating method first.
% \muhao{Need to say we diagnose these types of bias in benchmarks + crossref. But before we do that, we continue here by introducing the method for bias mitigation.}

\subsection{Counterfactual Data Augmentation}\label{sec:knowledge_conflict_mitigation}

% \muhao{should this be merged with 3, otherwise this part of the method descriptoin looks simple.}

\begin{figure}[t]
    \centering
    \includegraphics[width=\linewidth]{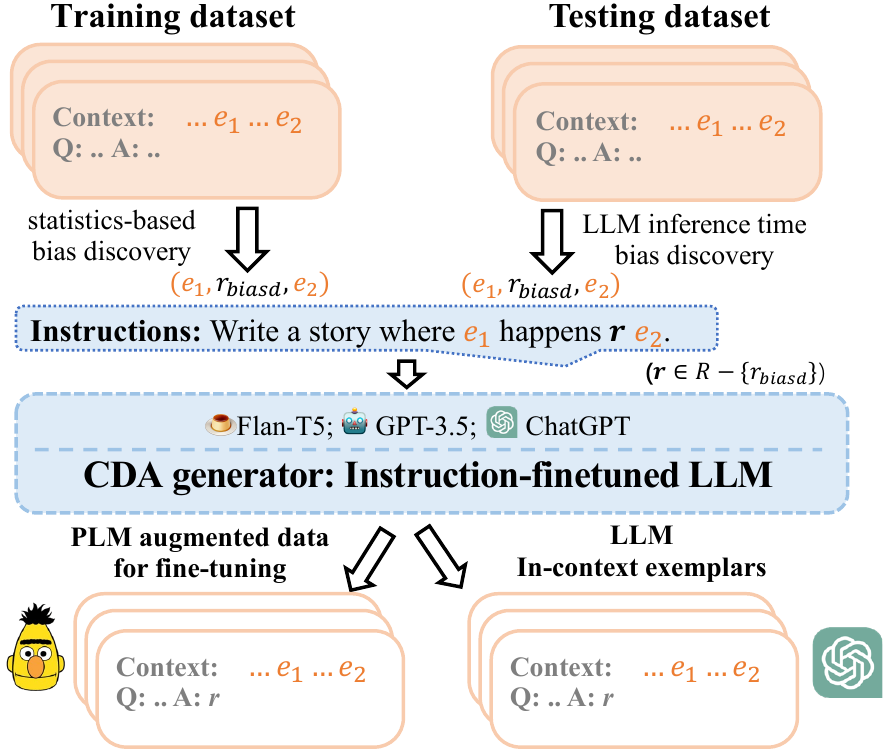}
    % \vspace{-0.1in}
    \caption{
    An overview of the CDA pipeline.
    }
    % \vspace{-1.5em}
    \label{fig:cda_overview}
\end{figure}

In this sub-section, we introduce our proposed Counterfactual Data Augmentation (CDA) method for mitigating knowledge conflicts (\Cref{fig:cda_overview}). We discuss the usage of CDA on both PLM and LLM separately, as they differ in their applications. Detailed adaptations and prompts will be introduced in \Cref{sec:exp_cda} and \Cref{sec:appendix_llm}.

\paragraph{Pre-trained Language Models.}
PLMs are usually fine-tuned on a training corpus, which naturally contains event-relation biases
that tend to be amplified after fine-tuning \cite{DBLP:journals/corr/abs-2201-11706}. 
% Further, those biases tend to be amplified after training on such a biased dataset \cite{DBLP:journals/corr/abs-2201-11706}. 
To mitigate bias, our proposed method automatically generates context that contains event pairs whose actual temporal relation is different from the biased relation. % that violate the bias patterns. 
Such knowledge-conflict (counterfactual) counterparts are trained together with the original training corpus to mitigate the biased training distribution.
To be more specific, for each event pair $(e_1, e_2)$ that is identified as biased, we ask an Instruction-finetuned Language Models~\cite{DBLP:journals/corr/abs-2210-11416} to generate context where $(e_1, e_2)$ %happens under a knowledge-conflict \textsc{TempRel}.
are associated with a \textsc{TempRel} that leads to a low bias score of a certain bias type.
Such augmented data can be regarded as knowledge-conflict data.
% , entitled augmented knowledge-conflict data.
The intuition is that, even though language models may suffer from bias and cannot directly solve the task, they can be well applied to generate synthetic data under structured instructions~\cite{DBLP:journals/corr/abs-2303-04132}.
% Detailed data augmentation prompts for each dataset we studied are presented in \Cref{sec:exp_setups}.

\paragraph{Large Language Models.}
The go-to way of using LLMs for downstream tasks is test-time In-Context Learning, as fine-tuning of the LLM is typically impractical or unviable.
In this case, we extend the idea of Counterfactual Data Augmentation to automatically generate counterfactual examples for in-context learning.
Unlike the data augmentation in PLMs, we generate counterfactual counterparts for every event pair to be studied.
% instead of only for the biased ones.
% Similar with the data augmentation paradigm in PLMs, 
For a new event pair $(e_1, e_2)$ to be studied, 
we first acquire the predicted relation $r_{LLM}$ by the LLM.
% which is regarded as a ``\textit{factual}'' prediction as it is what the LLM itself \textit{hallucinates}.
We leverage the LLM to generate context examples where $(e_1, e_2)$ are associated with relations that belong to $R - \{r_{LLM}\}$ as counterfactual examples to showcase the LLM the alternative cases when $(e_1, e_2)$ happens following a different \textsc{TempRel}. Note that this method is still considered a zero-shot as no training examples are seen during inference.

% Detailed data generation prompts for each dataset we studied are presented in Section \cref{sec:exp_setups}.

\section{Experiments}\label{sec:experiments}

% \muhao{Please add a lead paragraph.}
In this section, we introduce the datasets (\Cref{sec:dataset_intro}), the settings of knowledge conflict diagnosis (\Cref{sec:exp_conflict_diagnoses}), and conflict mitigation (\Cref{sec:exp_setups}), the primary experimental results and analysis (\Cref{sec:exp_results}).

\subsection{Datasets and Evaluation Metrics}\label{sec:dataset_intro}

We select two event temporal reasoning datasets\footnote{Details about datasets are presented in \Cref{sec:appendix_dataset}}.
% to study event-related knowledge conflicts:

\paragraph{TORQUE:}  
\citet{DBLP:conf/emnlp/NingWHPGR20torque} is a reading comprehension benchmark with a focus on event temporal reasoning questions. 
% TORQUE is more flexible than simple relation extraction benchmarks as the reading comprehension framework allows more complicated \textsc{TempRels} such as uncertain relations (e.g., \textit{might before}).
% including uncertain relations (e.g., \textit{might before}), hypothetical relations (e.g., \textit{what will happen if ...}), and negated relations (e.g., \textit{not after}).

% In TORQUE, each passage is associated with around 10 human-annotated questions regarding the \textsc{TempRel} between certain events, and the task objective is to select the correct answers from the pre-defined set of annotated event triggers.
% We evaluate the model performance using exact-match (EM) and Macro F1. 

\paragraph{MATRES:}
MATRES~\cite{DBLP:conf/acl/RothWN18matres} is a \textsc{TempRel} extraction dataset that includes refined annotations from documents in different domains.
% in TimeBank~\cite{pustejovsky2003timebank}, AQUAINT~\cite{louis2012corpus}, and Platinum~\cite{uzzaman2013semeval}. 
% The task in MATRES is defined as identifying the \textsc{TempRel} between two events in the context, where $R = \{\textit{before, after, equal, vague}\}$. 
The \textsc{TempRels} of interest are nailed to $R = \{\textit{before, after, equal, vague}\}$.
% We use the pre-processing by \citet{DBLP:conf/emnlp/WangCZR20} to process the dataset from raw annotations, where the context includes the sentences containing the two events $e_1$ and $e_2$, together with a precedent sentence to provide more contextual information.
We randomly sample 1,000 entries (out of $\sim$6k) from the development set to perform evaluations for LLMs\footnote{A common practice when doing GPT3-related experiments to reduce the overall cost~\cite{DBLP:journals/corr/abs-2303-16421}.}.

We use Exact-Match (EM) and F1 as the evaluation metric for TORQUE, and Micro/Macro F1 for MATRES. 
% Among them, 
(Macro) F1 is used as the primary metric, due to the imbalanced distribution of labels\footnote{Justifications are presented in \Cref{sec:appendix_eval}}.

\subsection{Knowledge Conflict Diagnosis}\label{sec:exp_conflict_diagnoses}

We apply the bias statistics introduced in \Cref{sec:knowledge_conflict_detection} on the training set to select knowledge-conflict subsets from both TORQUE and MATRES development sets. 
In MATRES, we directly make use of the \textsc{TempRel} information $(e_1, e_2, r)$ provided in each data entry to count the occurrence and calculate bias.
However, in TORQUE, the problem is formulated as reading comprehension, which requires further pre-processing to acquire pairwise \textsc{TempRels}.
Specifically, we parse each question to acquire the temporal predicate and arguments to form a $(e_1, e_2, r)$ format.
% For example, 
% for the question ``\textit{What happened after Bush gave four key speeches?}'' and answers ``\{\textit{called}, \textit{elect}, \textit{vote}\}'' under a certain context,
% we can acquire three event relation triples (\textit{gave}, \textit{before}, \textit{called}), (\textit{gave}, \textit{before}, \textit{elect}), and (\textit{gave}, \textit{before}, \textit{vote}).
% and use those triples for calculating and detecting bias.
In addition, TORQUE includes \textit{warm-up} questions that analyze whether a single event has \textit{happened}, \textit{will happen}, or \textit{is happening}. Our study calculates bias statistics based on a \textit{single} event and its temporal status 
% (\textit{happened}, \textit{will happen}, or \textit{is happening}) 
relative to a time expression in the context.
The bias in warm-up questions are labeled with \textit{warm-up}, while the other questions studying event-pair relations are labeled with \textit{relation}.

In addition, \Cref{tab:top_bias} lists the most biased features selected for both datasets.
We can find some 
bias, for example, a past tense is more often predicted as \textit{before} a present tense.
More details on the hyperparameters and statistics of knowledge conflict diagnosis are presented in \Cref{sec:appendix_bias_param}, \Cref{sec:appendix_knowledge_conflict_detection}, and \Cref{tab:bias_statistics} in the Appendix.
% The hyperparameters we used and analysis to sensitivity of $\mathcal{T}$ are listed in \Cref{sec:appendix_bias_param}.
% The statistics of the knowledge-conflict subsets we acquired are presented in \Cref{tab:bias_statistics} in the Appendix.

% For each type of bias, we empirically set thresholds to select knowledge-conflict subsets. 
% For a feature-relation pair $f$ (e.g., $f$ represents dependency) and $r$, it is knowledge-conflict if $b(f, r) < \frac{c(r)}{\sum_{r'\in R}c(r')}$, indicating that it does not conform to the dominant distribution of relation $r$.
% Such selection criteria can be further enhanced by setting a threshold $\mathcal{T}_r<\frac{c(r)}{\sum_{r'\in R}c(r')}$, which increases the level of conflicts by further restricting $b(f, r)$ to be less than $\mathcal{T}_r$.
% The hyperparameters we used are listed in \Cref{sec:appendix_bias_param}.
% The statistics of the knowledge-conflict subsets we acquired are presented in \Cref{tab:bias_statistics}.
% in the appendix.

\begin{table}[t]
\centering
\scriptsize
\renewcommand\arraystretch{1.1}
\newcolumntype{C}[1]{>{\centering\arraybackslash}p{#1}}
\begin{tabular}{l|p{22em}}
\toprule
  \multicolumn{1}{l}{\textbf{TORQUE}} &  \\
 \midrule
Rel.Prior & $b(\text{kill}, \text{arrest}, \textit{before})$=0.69, $b(\text{bombing}, \text{condemn}, \textit{before})$=0.67 \\\cline{2-2}
& $b(\text{incident}, \textit{happened})$=1,$b(\text{host}, \textit{future})$=0.91, $b(\text{progress}, \textit{happening})$=1\\
\hline
Tense & $b(\texttt{VBN}, \texttt{VB}, \textit{before})$=0.64,$b(\texttt{VBN}, \texttt{VBD}, \textit{before})$=0.48, $b(\texttt{VBD}, \texttt{VB}, \textit{before})$=0.55 \\ \cline{2-2}
& $b(\texttt{VBD}, \textit{happened})$=0.95,$b(\texttt{VB}, \textit{future})$=0.60, $b(\texttt{VBZ}, \textit{happening})$=0.62 \\
\hline
Narrative & $b(p1 \textit{<} p2, \textit{before})$=0.50,$b(p1 \textit{<} p2, \textit{after})$=0.32, $b(p1 \textit{<} p2, \textit{equal})$=0.03,$b(p1 \textit{<} p2, \textit{vague})$=0.13 \\
\hline
Dependency & $b(\texttt{xcomp}, \textit{before})$=0.81,$b(\texttt{ccomp}, \textit{after})$=0.70  \\
\bottomrule
\toprule
\multicolumn{1}{l}{\textbf{MATRES}} & \\
\midrule
Rel.Prior & $b(\text{say}, \text{have}, \textit{after})$=1,$b(\text{rise}, \text{close}, \textit{before})$=1, $b(\text{have}, \text{close}, \textit{before})$=0.83  \\
\hline
Tense & $b(\texttt{VBN}, \texttt{VB}, \textit{before})$=0.80,$b(\texttt{VBN}, \texttt{VBP}, \textit{before})$=0.78, $b(\texttt{VBD}, \texttt{VB}, \textit{before})$=0.70 \\
\hline
Narrative & $b(p1 \textit{<} p2, \textit{before})$=0.50,$b(p1 \textit{<} p2, \textit{after})$=0.32, $b(p1 \textit{<} p2, \textit{equal})$=0.03,$b(p1 \textit{<} p2, \textit{vague})$=0.13  \\
\hline
Dependency & $b(\texttt{xcomp}, \textit{before})$=0.61,$b(\texttt{ccomp}, \textit{after})$=0.60 \\
\bottomrule
\end{tabular}
\caption{Selected top biased event features in TORQUE and MATRES.  }
\label{tab:top_bias}
\end{table}

\begin{table*}[t]
\centering
\small
\renewcommand\arraystretch{1.0}
\newcolumntype{C}[1]{>{\centering\arraybackslash}p{#1}}
\begin{tabular}{p{5.0em}|C{1.2em}C{1.2em}|C{1.4em}C{1.4em}C{1.4em}C{1.4em}C{1.4em}C{1.4em}C{1.4em}C{1.4em}C{1.4em}C{1.4em}C{1.4em}C{1.4em}|C{1.2em}C{1.2em}}
\toprule
 & \multicolumn{2}{c|}{\textbf{all}} & \multicolumn{2}{c}{ \tabincell{c}{\textbf{Rel.Prior}\\ (relation)} } & \multicolumn{2}{c}{\tabincell{c}{\textbf{Rel.Prior}\\(warm-up)}} & \multicolumn{2}{c}{\tabincell{c}{\textbf{Narrative}\\ (relation)}} & \multicolumn{2}{c}{\tabincell{c}{\textbf{Tense}\\(relation)}} & \multicolumn{2}{c}{\tabincell{c}{\textbf{Tense}\\(warm-up)}} & \multicolumn{2}{c|}{\tabincell{c}{\textbf{Dep.}\\(relation)}} & \multicolumn{2}{c}{\tabincell{c}{\textbf{Confl.Avg.}}} \\
 \cline{2-17}
 & EM & F1$^*$ & EM & F1$^*$ & EM & F1$^*$ & EM & F1$^*$ & EM & F1$^*$ & EM & F1$^*$ & EM & F1$^*$ & EM & F1$^*$  \\
\midrule
\multicolumn{1}{l}{\textbf{PLM}} \\
\hline
RoBERTa-L & 50.4 & 75.7 & {29.5} & 	73.3& 	50.0& 	\underline{75.1}& 	\underline{31.4}& 	\underline{69.0}& 	33.5& 	72.9& 	48.4& 	\underline{72.4}& 	41.7& 	78.6 & 39.1 & \underline{73.6}    \\
PoE & 33.3 &65.8 & 21.6 & \textbf{76.1} & 22.7 & 59.8 & 23.5 &67.1&27.5&71.1&22.5&57.0&32.3&\textbf{79.2} & 25.0 & 68.4 \\
L.-mixin &46.8&74.8&27.2&75.2&50.0&72.1&27.8&68.4&30.8&72.6&\textbf{49.3}&69.8&33.8&76.8 & 36.5 & 72.5 \\
L.-mixin+H & 37.6 & 70.6&20.4&73.4&40.9&71.6&28.5&\textbf{69.6}&28.8&71.6&38.0&67.7&32.3&76.0 & 31.5 & 71.7  \\
Cont. Inf. & \textbf{53.1} & \underline{75.9} & 28.4 & 75.3 & \underline{50.0} & 72.5 & \textbf{35.7} & 68.9 & \underline{35.4} & \textbf{73.1} & \textbf{49.3} & 70.2 & \textbf{44.1} & 78.9 & \underline{40.5} & 73.2 \\
% Contrastive \\
AFLite&50.5&75.8&\textbf{34.1}&73.5&48.5&72.1&26.4&68.2&34.6&72.7&47.9&69.8&39.7&77.3&38.5&72.3\\
CDA (Ours)  & \underline{51.0}&\textbf{76.1}&\underline{33.7}&\underline{75.4}&\textbf{50.0}&\textbf{75.9}&30.7&68.6&\textbf{35.5}&\textbf{73.1}&{48.8}&\textbf{73.2}&\textbf{44.1}&\underline{79.1} & \textbf{40.5} & \textbf{74.2} \\

\midrule
\multicolumn{1}{l}{\textbf{LLM}} \\
\hline
GPT-3.5 & \textbf{8.36} & \textbf{45.5} & 4.82 & 59.9 & \underline{4.62} & 47.0 & 2.13 & 50.7 & 4.46 & 53.5 & \underline{5.71} & 45.9 & 2.94 & 57.7 & \underline{4.12} & 52.5 \\
\quad + ICL & \underline{7.22} & {44.9} & \textbf{9.09} & \underline{60.2} & \textbf{9.09} & \textbf{55.6} & \underline{2.14} & 51.3 & \textbf{5.35} & \underline{55.5} &  \textbf{8.45} & \textbf{52.6} &  \textbf{4.41} & \underline{58.8} & \textbf{6.42} & \textbf{55.7} \\
% \quad w/ KATE \\
\quad + GDA & 4.85 & 44.0 &  \underline{5.68} & 60.0 & 1.54 & \underline{49.4} & \textbf{3.19} & \underline{54.6} & 3.18 & \textbf{56.1} & 1.43 & {48.3} & 2.94 & 58.6 & 3.00 & 54.5 \\
% \quad + CDA & 5.16 & 44.6 &  4.55 & \textbf{60.5} & 3.08 & \underline{49.5} & \textbf{3.19} & \textbf{54.6} & \textbf{5.73} & \textbf{56.9} & 2.86 & 47.8 &  \textbf{5.88} & \textbf{59.7} & \underline{4.22} & \underline{54.9}  \\
\quad + CDA & 5.53 & \underline{45.1} & \underline{5.68} & \textbf{60.6} & 1.52  & 48.0 & \underline{2.14} & \textbf{56.5} & \underline{4.53} & 54.1 & 1.41 & \underline{50.1} & 2.94 & \textbf{61.2} & 3.04 & \underline{55.1} \\
\hline
ChatGPT & \textbf{17.7} & 40.7 & \textbf{9.09} & 40.3 & \textbf{4.55} & 38.3 & \textbf{6.43} & 42.3 & \textbf{10.3} & 41.4 & \textbf{4.23} & 35.8 & \textbf{7.35} & 42.2 & \textbf{6.99} & 40.0  \\
\quad + ICL & 3.92 & 43.9 & \underline{4.55} & \textbf{58.3} & \textbf{4.55} & 50.1 & 1.43 & 48.9 & \underline{3.70} & 52.8 & \textbf{4.23} & 47.9 & 1.47 & 54.8 & \underline{3.32} & 52.1 \\
\quad + GDA &  4.38 & \underline{44.2} & 3.41 & \underline{56.2} & 1.52 & \underline{50.6} & 1.43 & \underline{50.0} & 3.29 & \underline{52.9} & 1.41 & \underline{48.3} & 2.94 & \underline{57.4} & 2.33 & \underline{52.6} \\
\quad + CDA & \underline{6.72} & \textbf{45.2} & 3.41 & 55.6 & 1.52 & \textbf{50.9} & 1.43 & \textbf{51.4} & 2.06 & \textbf{53.3} & 2.82 & \textbf{50.0} & \textbf{4.41} & \textbf{59.1} & 2.60 & \textbf{53.3} \\

\bottomrule
\end{tabular}
% \vspace{-0.1in}
\caption{Experimental results on the TORQUE dataset. Exact-Match (EM) rate and Macro-F1 (F1, regarded as the primary metric $^*$ since EM can be susceptible to manipulation by simply predicting `none') scores are reported. 
Best-performed results are \textbf{bold-faced} and the second-best are \underline{underlined}. 
% LLM + CDA outperforms its directly comparable baseline GDA.
}
\label{tab:torque_main_results}
% \vspace{-0.1in}
\end{table*}

\subsection{Setup for Conflict Mitigation}\label{sec:exp_setups}

\paragraph{Counterfactual Data Augmentation.}\label{sec:exp_cda}
We introduce the details of conducting Counterfactual Data Augmentation here. 
In augmentations for PLM, we choose Flan-T5 (11B)~\cite{DBLP:journals/corr/abs-2210-11416} as a more scalable generator (than API-based LLMs).
For each event pair $(e_1, e_2, r)$ identified as being biased according to \textit{Relation Prior Bias}, we generate context with the prompt \textit{Write a story where $e_1$ happens $r'$ $e_2$:}, where $r' \in R - \{r\}$ (e.g., $r'$=\textit{before}).
In TORQUE, we thus construct a question $Q$=\textit{``What happened $r'$ $e_2$''}, and the corresponding answer is $e_1$.
% In MATRES, we require the model to directly predict the relation $r$ given the generated context.
Based on the coarse data, we apply additional filters  
% We retain only the generated data that conflict with the biased distributions so that the relational distribution of various features can be more balanced.
to only retain those that are not biased in terms of tense and narrative.

For LLMs, we ask the model itself to predict the labels of the test data first.
Take MATRES as an example, denote $r_{LLM}$ as the \textit{factual} prediction by the LLM, and then we ask the LLM itself to \textit{Generate a paragraph where event $e_1$ happens $r'$ $e_2$}, where $r'\in R-\{r_{LLM}\}$.
More detailed prompts are presented in \Cref{sec:appendix_llm}.

\paragraph{Model Configuration.} We perform experiments using both PLMs and LLMs
% \footnote{We refer readers to \Cref{sec:appendix_model} for more details experimental setups. We also present the effects of different prompt templates and the number of few-shot exemplars.}
.
% To use PLM for TORQUE, we take the concatenation of the context and the question as inputs to the PLM.
% The model has a one-layer MLP on top of the encoder and the input to the MLP is the transformers' output corresponding to all the event triggers in the context.
% The training objective is defined to predict the correct answers among all the annotated event triggers.
% For MATRES, we take the transformers' output embedding of the two target events to feed into a two-layer MLP to predict the relations out of four choices. 
% We use RoBERTa-large as the backbone model as in the original papers. 
% We use a batch size of 6 and learning rate $1e^{-5}$ for TORQUE and batch size of 64 and learning rate  $1e^{-5}$ for MATRES.
% To use LLMs, we use the following prompt template for TORQUE: \textit{Q: [question], select none or several from [all events]} \textit{[context] $\backslash$n A:}.
For PLMs we use RoBERTa-large~\cite{DBLP:journals/corr/abs-1907-11692} as the backbone.
We use GPT-3.5 (\texttt{text-davinci-003}) and ChatGPT (\texttt{gpt-3.5-turbo}) as the backbone LLM\footnote{Details of prompts are listed in \Cref{sec:appendix_llm}.}. 
% We parse the answers to acquire the predictions.
% For MATRES, 
% we formalize the problem as a multi-choice question-answering format.
% \textit{Given the context:$\backslash$n, [context], $\backslash$n$\backslash$n Q: What's the \textsc{TempRel} between the event [e1] and [e2]? $\backslash$n Choice A: [e1] happens before [e2]. $\backslash$n Choice B: [e1] happens after [e2]. $\backslash$n Choice C: [e1] happens during [e2]. $\backslash$n Choice D: unknown. $\backslash$n Answer only with A, B, C, or D. $\backslash$n$\backslash$n A: Choice"}. 

\paragraph{Baselines.}
We compare our proposed methods with other representative bias mitigation approaches, including Product-of-Experts (PoE; \citealt{DBLP:journals/neco/Hinton02,he-etal-2019-unlearn}), Learned-mixin~\cite{DBLP:conf/emnlp/ClarkYZ19}, Counterfactual Inference~\cite{DBLP:conf/naacl/WangCZCLLYLH22, DBLP:conf/eacl/WangZDGRC23}, and AFLite~\cite{le2020adversarial}. 
These baselines are typical bias-agnostic debiasing baselines that address %unknown bias
known or unknown bias with statistical approaches. 
For LLMs, we use the vanilla In-Context Learning (ICL) by randomly retrieving one set of exemplars from the training set as demonstrations. 
% Specifically, we retrieve one passage-question pair in TORQUE, and retrieve one example per relation from \textit{before}, \textit{after}, \textit{equal}, and \textit{unknown}, to form the ICL demonstration.
Note that ICL is considered few-shot learning while our method is purely \textit{zero-shot}.
In addition, to study the effect of the strategy for generating \textit{counterfactual} exemplars, we add an additional baseline named \textit{Generative Data Augmentation} (GDA) that performs exemplar generation without counterfactual guidance\footnote{Details of all baselines are in \Cref{sec:appendix_baseline}}. 
% That is to say, we ask LLMs to generate exemplars under all relations from $R$, instead of only the counterfactual relations. 

\begin{table*}[t]
\centering
\small
\renewcommand\arraystretch{1.0}
\newcolumntype{C}[1]{>{\centering\arraybackslash}p{#1}}
\begin{tabular}{l|C{2.1em}C{2.1em}|C{2.1em}C{2.1em}C{2.1em}C{2.1em}C{2.1em}C{2.1em}C{2.1em}C{2.1em}|C{2.1em}C{2.1em}}
\toprule
 & \multicolumn{2}{c|}{\textbf{all}} & \multicolumn{2}{c}{ \textbf{Rel. Prior} }  & \multicolumn{2}{c}{\textbf{Narrative}} & \multicolumn{2}{c}{\textbf{Tense}} & \multicolumn{2}{c|}{\textbf{Dependency}} & \multicolumn{2}{c}{\textbf{Confl.Avg.}} \\
 \cline{2-13}
 & \footnotesize{Micro} & \multicolumn{1}{@{}p{2.1em}|}{\footnotesize{Macro}$^*$}  & \footnotesize{Micro} & \footnotesize{Macro}$^*$ & \footnotesize{Micro} & \footnotesize{Macro}$^*$ & \footnotesize{Micro} & \footnotesize{Macro}$^*$ & \footnotesize{Micro} & \multicolumn{1}{@{}l|}{\footnotesize{Macro}$^*$} & \footnotesize{Micro} & \multicolumn{1}{@{}p{2.1em}}{\footnotesize{Macro}$^*$}  \\
\midrule
\multicolumn{1}{l}{\textbf{PLM}} \\
\hline
RoBERTa-large &  \underline{70.8} & 44.9 & 59.7 & 28.5 &   \underline{59.2} & 27.1 & 54.8 & \underline{33.2} &  58.5 & 38.3 & 58.0 & 31.8 \\
PoE % &68.8&46.0&60.1&31.6&51.2&33.8&62.4&29.9&57.7&40.1\\
% SD=41&69.4&46.7&60.1&32.7&57.7&40.4&58.9&29.4&60.0&43.0\\
&69.4&45.3&60.0&\underline{30.7}&52.6&32.8&61.1&29.0&53.1&36.7 & 56.7 & 32.3\\
Learned-mixin&71.0&45.0&\underline{60.4}&29.5&55.7&\textbf{34.6}&60.9&27.5&\underline{60.0}&\textbf{40.1} & \underline{59.2} & \underline{32.9} \\
% SD=41&69.4&46.0&61.8&31.7&54.2&35.9&59.5&28.7&63.1&43.5\\
% SD=42&70.5&46.4&65.3&33.4&55.1&35.1&57.0&28.3&53.8&37.7\\
Learned-mixin+H&70.5&44.8&59.6&29.2&54.3&\underline{34.0}&\underline{62.2}&27.7&58.5&39.8 & 58.6 & 32.6 \\
% SD=41&68.8&46.2&61.5&32.2&53.6&36.3&59.4&29.1&61.2&42.6\\
% SD=42&71.5&46.4&63.0&32.2&57.6&36.8&57.8&28.1&56.5&37.7\\
Cont. Inf. & 67.6 & \underline{45.0} & 60.3 & \textbf{31.4} &  \textbf{60.7} & 27.3 &  48.8 &  32.5 & 55.3 & 38.9 & 56.3 & 32.5 \\
% Contrastive \\
AFLite&64.3&43.4&52.4&28.8&50.3&32.8&\textbf{62.5}&30.0&55.0&39.3&55.1&32.7\\
CDA (Ours) &  \textbf{72.2} & \textbf{45.5} & \textbf{61.5} & {29.3} &  58.8 & {27.3} & {57.2} & \textbf{35.1} &  \textbf{62.2} & \underline{39.9} & \textbf{59.9} & \textbf{32.9} \\
\midrule
\multicolumn{1}{l}{\textbf{LLM}} \\
\hline
GPT-3.5 &   \textbf{53.3} & 19.7 & \underline{54.7} & 25.3 & 2.57 & 3.98 & 36.7 & 17.2 & 28.6 & 13.0 & 30.6 & 14.9 \\
\quad + ICL & \underline{51.6} & 18.4 &  \textbf{56.1} & 20.9 & 1.52 & 2.31 & 35.7 & 16.4 & 26.2 & 10.6 & 29.9 & 12.6 \\
\quad + GDA & 45.6 & \underline{27.6} &  52.0 & \underline{32.4} & \underline{15.1} & \underline{14.9}  & \underline{37.6} & \underline{24.0} & \textbf{33.3} & \underline{18.9} & \underline{34.5} & \underline{22.6} \\
\quad + CDA &  51.3 & \textbf{30.0} & 53.4 & \textbf{36.0} & \textbf{16.6} & \textbf{26.8} & \textbf{38.1} & \textbf{27.2} &  \textbf{33.3} & \textbf{21.5} & \textbf{35.4} & \textbf{27.9} \\
\hline
ChatGPT  & 39.8 & 25.9 & 31.1 & 22.3 & \textbf{37.6} & \textbf{32.5} & 27.0 & 17.6 & 21.4 & 13.8 & 29.3 & 21.6 \\
\quad + ICL & 43.1 & 23.8 & \textbf{53.4} & 23.5 & 34.8 & 22.2 & 11.3 & 12.7 & 28.6 & 11.1 & 32.0 & 17.4 \\
\quad + GDA  & 45.7 & 30.8 & 36.5 & \textbf{25.1} & 29.5 & 26.2 & \textbf{32.5} & \textbf{20.7} & \textbf{40.5} & \textbf{24.4} & \underline{34.7} & \textbf{24.1} \\
\quad + CDA & \textbf{49.3} & \textbf{32.0} & \underline{42.6} & \underline{24.3} & \underline{37.1} & \underline{31.0} & \underline{31.2} & \textbf{20.7} & \underline{33.3} & \underline{19.3} & \textbf{36.1} & \underline{23.8} \\
\bottomrule 
\end{tabular}
% \vspace{-0.1in}
\caption{Experimental results on MATRES. We use two evaluation metrics, Micro-F1 (denoted as Micro) and  Macro F1 (denoted as Macro; regarded as the primary metric $^*$ due to the significant class imbalance). Best-performed results are \textbf{bold-faced} and the second-best is \underline{underlined}.}
\label{tab:matres_main_results}
% \vspace{-0.1in}
\end{table*}

\subsection{Results and Analysis}\label{sec:exp_results}

We present the main experimental results for TORQUE in \Cref{tab:torque_main_results} and for MATRES in \Cref{tab:matres_main_results}\footnote{As elaborated in \Cref{sec:appendix_dataset}, we use a different preprocessing of MATRES that includes an additional context sentence, making the performance different than \citet{DBLP:conf/acl/RothWN18matres}}.
% The performance of PLMs and LLMs are separately compared in different rows. 
% We also include the average of all bias categories in the last column.
The \textit{all} row indicates the performance on the whole evaluation set.
The \textit{Confl.Avg.} column is an average of all knowledge-conflict subsets, measuring models' ability on mitigating knowledge conflicts.
The columns in between indicate the performance on each knowledge conflict types, evaluated on the detected subsets.

\paragraph{Impact of Knowledge Conflicts.}
Models on both TORQUE and MATRES show a decrease in performance when evaluated on knowledge-conflict subsets. 
\Cref{tab:effect_conflict_matres} shows a comparison of baseline model performance on the conflict and non-conflict partitions of MATRES.
The comparison on TORQUE is presented in \Cref{tab:effect_conflict} in the Appendix, showing a similar trend.
This finding indicates that the selected conflict subsets are indeed more confusing for language models, proving the effectiveness of our conflict detection framework.
% This can be attributed to the fact that events connected by a dependency relation are easier to classify. 
% Entries where events have an explicit dependency relation (including biased and unbiased) yield EM and Macro-F1 scores of 43.0\% and 79.8\%, respectively, indicating a performance drop in the models on knowledge-conflict subsets with regard to dependency bias.

For LLMs, the overall performance is not satisfactory compared with fully-supervised models, which is in line with the findings in several evaluation works on LLMs \cite{DBLP:journals/corr/abs-2304-14827, zhou2023context, DBLP:journals/corr/abs-2304-05454}, 
due to the fact that such tasks focusing on specific types of contextualized reasoning, when not trained with instruction fine-tuning,  
often lead to poor performance~\cite{zhang2023aligning}. 
% We present more analysis on prompt designs for the two tasks in the Appendix~\ref{sec:appendix_llm_prompt}.
Nonetheless, since LLMs are not fine-tuned on the biased training set, their performance on knowledge-conflict subsets %is not significantly worse than their performance
does not drop as significantly in comparison to that on the entire evaluation set, %In fact, in some cases, their performance on knowledge-conflict subsets is even better than that on the entire evaluation set. 
while even being better in some cases.
This suggests that zero-shot predictions using LLM can be more generalizable
% equitable 
when not trained on smaller and biased data.

% Another noteworthy point is the inconsistent trends between LLM performance regarding EM/F1 in TORQUE and Macro/Micro F1. 
% This is due to the imbalanced label distribution in both datasets, and the tendency of LLMs to predict the most dominant label, which is ``None'' in TORQUE and ``before'' in MATRES. More details are presented in \Cref{sec:appendix_eval}.

\begin{table}[t]
\centering
\small
\renewcommand\arraystretch{1.0}
\newcolumntype{C}[1]{>{\centering\arraybackslash}p{#1}}
\begin{tabular}{l|cccc}
\toprule
 & \multicolumn{2}{c}{Conflict} & \multicolumn{2}{c}{Non-Conflict} \\
 \cline{2-5}
 & Micro & Macro & Micro & Macro \\
\midrule
\multicolumn{2}{l}{\textbf{RoBERTa-large}} \\ 
\hline
Relation Prior & 59.7$\downarrow$ & 28.5$\downarrow$ & 75.7 & 40.9 \\
Narrative & 59.2$\downarrow$ & 27.1$\uparrow$ & 76.8 & 21.7 \\
Tense & 54.8$\downarrow$ & 33.2$\downarrow$ & 72.8 & 47.2 \\
Dependency & 58.5$\downarrow$ & 38.3$\downarrow$ & 70.0 & 45.7 \\
\midrule
\multicolumn{2}{l}{\textbf{GPT-3.5}} \\ 
\hline
Relation Prior & 54.7$\downarrow$ & 25.3$\downarrow$ & 56.8 & 28.6 \\
Narrative & 2.57$\downarrow$ & 3.98$\downarrow$ & 85.8 & 26.3 \\
Tense & 36.7$\downarrow$ & 17.2$\downarrow$ & 60.3 & 27.2 \\
Dependency  & 28.6$\downarrow$ & 13.0$\downarrow$ & 57.7 & 28.9 \\
\bottomrule
\end{tabular}
% \vspace{-0.1in}
\caption{Performance on knowledge conflict and non-conflict data in MATRES. Both models suffer from a performance drop when tested on the conflict subsets. $\downarrow$ indicates a performance drop in the conflict subsets.}
\label{tab:effect_conflict_matres}
% \vspace{-1em}
\end{table}

\paragraph{Knowledge Conflicts Mitigation.}
CDA significantly improves the performance of the  vanilla PLM RoBERTa-large both on the entire evaluation set and on each of the knowledge-conflict subsets. 
Bias-agnostic baselines adopt a model trained only with event arguments and without context, which performs debiasing by countering event-relation bias. 
This yields competent results related to the relationship prior bias. 
The counterfactual inference is more effective than other fine-tuned-based methods, as also reported by previous work~\cite{DBLP:conf/naacl/WangCZCLLYLH22}. 
However, bias-aware data augmentation methods are generally more effective, as they explicitly address different forms of bias and have a more focused performance on biased datasets.

As for LLMs, on MATRES, CDA-based demonstrators can improve the performance on both the whole evaluation set and all the knowledge conflict datasets, with the exception of a minor setback compared to ChatGPT-GDA in terms of \textit{Confl.Avg.} Macro-F1.
% On TORQUE, the paradigm for generating demonstrators, either GDA or CDA, cannot %beat 
% outperform standard ICL that retrieves random exemplars from the training set. 
% This may be due to the fact that the data entries in TORQUE are more complicated and include more events and relations than those in MATRES.
On TORQUE, CDA on ChatGPT outperforms all baselines in terms of overall performance and \textit{Confl.Avg.} on the main metric F1. 
For GPT-3.5, the zero-shot setting surprisingly achieves the best overall performance. 
However, CDA can outperform GDA, indicating that adding a counterfactual prior can better help LLMs to understand event temporal reasoning.
Another noteworthy point is that our CDA method is purely zero-shot compared with ICL, showing the superiority of applying counterfactual guidance to LLMs.

\paragraph{Error Analysis on Different Bias Types}

Taking MATRES as an example, zero-shot prediction by GPT-3.5 suffers significantly from narrative bias, where the performance is near-zero. This finding is consistent with two other different prompt templates (\Cref{tab:matres_prompt_analysis} in the appendix). This is mainly because GPT is an autoregressive model following a single-directional encoding, and it may not be fine-tuned on temporal reasoning data to understand the chronological orders, making it uses a shortcut to rely on positions of events to conduct reasoning.
ChatGPT, on the other hand, suffers from dependency bias the most, where there are syntactic dependencies between the two target events. This may be attributed to that ChatGPT, though having a stronger text generation ability, falls short of solving more subtle and contextual cases for temporal reasoning, as events that have a direct dependency edge typically occur in close proximity within the context. However, reasoning about such scenarios can be challenging, even for humans, due to their subtle nature~\cite{DBLP:journals/corr/abs-2304-05454}.

\paragraph{Quality Analysis of Generated Data}
As the generated data are used for better training/prompting language models, the quality of them is reflected by the downstream performance improvement. 
In the appendix, we compare our CDA with popular task-agnostic data augmentation techniques in \Cref{tab:torque_data_aug} and \Cref{tab:matres_data_aug}, and show that our CDA method can better help boost both the performance in terms of both knowledge-conflict and overall performance.
Regarding LLMs, the comparison between GDA and our CDA also demonstrated the fact the augmented data with a counterfactual constraint can better help both overall and knowledge-conflict reasoning ability.

\paragraph{Why CDA doesn't outperform ICL in TORQUE.} 
For GPT-3.5, ICL indeed consistently outperforms our proposed CDA. We have manually checked the plausibility of the generated exemplars by CDA on GPT-3.5, and find that 4 out of 10 generated exemplars are either incomplete, or do not fully contain the events we desired. In this sense, the exemplars sampled from the training set are of a better quality than GPT-3.5-generated ones, leading to better ICL performance. For ChatGPT, on the other hand, the quality of the generated synthetic data is of a significantly better quality than GPT-3.5, resulting in more improvement on top of zero-shot ChatGPT and one-shot ICL for ChatGPT.

In all, the result of CDA on LLMs is highly dependent on 1) the data synthesizing capability of the backbone LLM, where ChatGPT excels, and 2) the problem solving ability for temporal reasoning, where GPT-3.5 excels. We also checked using the exemplars generated by ChatGPT as in-context examples for GPT-3.5, and the Confl.Avg. is 7.23 and 56.2, which is better than GPT-3.5-ICL. 

\paragraph{Case Study}
For example, for a typical tense-biased relation in \Cref{tab:bias_examples}, the ``told-offer’’ case, where past tense should happen after the present tense. We ask ChatGPT to generate the case when told happen after the present tense as a counterfactual exemplar: ``\textit{I offer my friend a ride to the party all the time. She told me she has already made plans to go with someone else. I understand and told her to let me know if she needs a ride in the future.}''
This gives a case when \textit{told} happens after \textit{offer} and can be used as a counterfactual augmented data. With this counterfactual exempler, LLMs can perform correctly in this case.

We also provide some additional analysis regarding why CDA cannot outperform ICL in TORQUE for GPT3.5 and the inconsistent trends of primary and secondary metrics on LLMs in \Cref{sec:appendix_analysis}.

\section{Conclusion}
In this paper, we investigate knowledge conflicts in event temporal reasoning by formally defining four types of biases to identify a knowledge conflict diagnoses evaluation set. We observe that both PLMs and LLMs are susceptible to knowledge conflicts in this task, resulting in decreased performance on knowledge-conflict datasets. 
To address this issue, we propose a CDA method that is suitable for both PLMs through pre-training and LLMs through In-Context Learning. 
Our experiments demonstrate the effectiveness of our proposed method in mitigating knowledge conflicts.

\section*{Acknowledgement}
Tianqing Fang was supported by the Hong Kong PhD Fellowship Scheme.
Wenxuan Zhou and Muhao Chen were supported by the NSF Grants IIS 2105329 and ITE 2333736, an Amazon Research Award and a Cisco Research Award.
Yangqiu Song was supported by the NSFC Fund (U20B2053) from the NSFC of China, the RIF (R6020-19 and R6021-20) and the GRF (16211520 and 16205322) from RGC of Hong Kong. 
Yangqiu Song thanks the support from the UGC Research Matching Grants (RMGS20EG01-D, RMGS20CR11, RMGS20CR12, RMGS20EG19, RMGS20EG21, RMGS23CR05, RMGS23EG08).
Tianqing Fang and Yangqiu Song also thank the support from Tencent AI lab.

\section*{Limitations}

This paper only discussed bias calculated based on statistics in the training set.
However, there are various other ways of characterizing bias, such as using predictions of zero-shot pre-trained language models~\cite{DBLP:conf/emnlp/XuWLDC22} and context masking, are not discussed, which can be left as a future work.

\section*{Ethics Statement}

There are no direct societal implications of this work. 
The datasets we use, TORQUE and MATRES, are publicly available and shared via open-access licenses for research purposes.
Even though we are detecting bias and conflicts in the original datasets, we focus on bias toward temporal relations of events and do not involve any bias toward certain gender or ethnics groups.
The context where event relations are derived from TimeBank\footnote{https://catalog.ldc.upenn.edu/LDC2006T08}, AQUAINT\footnote{https://catalog.ldc.upenn.edu/LDC2002T31}, and Platinum\footnote{https://bitbucket.org/leondz/te3-platinum}, which has not shown to contain any obvious social biases that would raise concerns within the community.
The counterfactual data augmentation technique we propose can effectively mitigate bias in event relation extraction.
In conclusion, to the best of our knowledge, this paper does not raise ethical concerns.

\bibliography{custom}

\appendix

\clearpage

\noindent
{\Large\textbf{Appendices}}

\section{Knowledge-conflict Selection Hyperparameters}\label{sec:appendix_bias_param}

In TORQUE, we set an empirical $\mathcal{T}_{\textit{before}}^{\text{Relation Prior}} = \mathcal{T}_{\textit{after}}^{\text{Relation Prior}} = \mathcal{T}_{\textit{equal}}^{\text{Relation Prior}} = 0.25$ by investigating the distribution of \textit{before}, \textit{after}, and \textit{equal} relations. 
For tense bias, we set $\mathcal{T}_{\textit{before}}^{\text{tense}} = \mathcal{T}_{\textit{after}}^{\text{tense}} = 0.25$, and
$\mathcal{T}_{\textit{equal}}^{\text{tense}} = 0.2$ for the relations indicating two events happening simultaneously. 
For narrative and dependency bias, the threshold is simply set as 0.5.
In MATRES, we set $\mathcal{T}_{\textit{before}} = \mathcal{T}_{\textit{after}} = 0.3$ and $\mathcal{T}_{\textit{equal}}=0.1$.

\paragraph{Sensitivity Analysis}
In MATRES, for example, if the $\mathcal{T}$ is set as its upper bound, 
$\mathcal{T}_{\textit{before}}=0.523$, and 
$\mathcal{T}_{\textit{after}}=0.346$, then more knowledge-conflict examples will be selected based on the new threshold. 
In MATRES, the number of Rel.Prior increases from 148 to 229, the number of tense prior increases from 210 to 418, and dependency prior from 42 to 69. The statistics of narrative bias don't change because the order of two events in the context is a binary relation and will not be affected by Tau. We reproduce the results in Table~\ref{tab:effect_conflict_threshold_matres} under the new $\mathcal{T}$, which studies the model performance on the conflict subset versus the non-conflict subset. 
The results are basically consistent with the $\mathcal{T}$ that is originally used in this paper, which shows that the PLM RoBERTa-large suffer from knowledge conflicts with a performance drop on the conflict dataset.

\begin{table}[h]
\centering
\small
\renewcommand\arraystretch{1.0}
\newcolumntype{C}[1]{>{\centering\arraybackslash}p{#1}}
\begin{tabular}{l|cccc}
\toprule
 & \multicolumn{2}{c}{Conflict} & \multicolumn{2}{c}{Non-Conflict} \\
 \cline{2-5}
 & Micro & Macro & Micro & Macro \\
\midrule
\multicolumn{2}{l}{\textbf{RoBERTa-large}} \\ 
\hline
Relation Prior & 62.7$\downarrow$ &	30.8$\downarrow$ &	75.4	& 40.1 \\
Narrative &  59.2$\downarrow$	&27.1$\uparrow$&	76.8	&21.7\\
Tense & 59.9$\downarrow$	&36.1$\downarrow$	&71.8	&46.0 \\
Dependency & 57.2$\downarrow$	& 37.8$\downarrow$	& 71.1	& 44.7 \\
\bottomrule
\end{tabular}
\vspace{-0.1in}
\caption{Performance on knowledge conflict and non-conflict data in MATRES under the new $\mathcal{T}$.}
\label{tab:effect_conflict_threshold_matres}
\vspace{-1em}
\end{table}

\section{Datasets}\label{sec:appendix_dataset}

\paragraph{TORQUE:}  
\citet{DBLP:conf/emnlp/NingWHPGR20torque} is a reading comprehension benchmark with a focus on event temporal reasoning questions. 
TORQUE is more flexible than simple relation extraction benchmarks as the reading comprehension framework allows more complicated \textsc{TempRels} 
% such as uncertain relations (e.g., \textit{might before}).
including uncertain relations (e.g., \textit{might before}), hypothetical relations (e.g., \textit{what will happen if ...}), and negated relations (e.g., \textit{not after}).
In TORQUE, each passage is associated with around 10 human-annotated questions regarding the \textsc{TempRel} between certain events, and the task objective is to select the correct answers from the pre-defined set of annotated event triggers.
We evaluate the model performance using exact-match (EM) and Macro F1.

\paragraph{MATRES:}
MATRES~\cite{DBLP:conf/acl/RothWN18matres} is a \textsc{TempRel} extraction dataset that includes refined annotations from documents
in TimeBank~\cite{pustejovsky2003timebank}, AQUAINT~\cite{louis2012corpus}, and Platinum~\cite{uzzaman2013semeval}. 
The task in MATRES is defined as identifying the \textsc{TempRel} between two events in the context, where $R = \{\textit{before, after, equal, vague}\}$. 
The \textsc{TempRels} of interest are nailed to $R = \{\textit{before, after, equal, vague}\}$.
We use the pre-processing by \citet{DBLP:conf/emnlp/WangCZR20} to process the dataset from raw annotations, where the context includes the sentences containing the two events $e_1$ and $e_2$, together with a precedent sentence to provide more contextual information.
This makes the context a bit longer than the common preprocessing adopted by several previous works such as \citet{DBLP:conf/acl/RothWN18matres, DBLP:conf/eacl/WangZDGRC23}, accounting for the discrepancy of model performances.

% We randomly sample 1,000 entries (out of $\sim$6k) from the development set to perform evaluations for LLMs\footnote{A common practice when doing GPT3-related experiments to reduce the overall cost~\cite{DBLP:journals/corr/abs-2303-16421}.}.

\section{Evaluation Metrics}\label{sec:appendix_eval}

In this paper, (binary) F1 is the primary metric in TORQUE where EM is the secondary metric. In MATRES, the primary metric is macro F1 and the secondary metric is micro F1.

\paragraph{Explanations on the Primary Metrics} In TORQUE, around 22\% of the evaluation entries have no answers, which means we can easily achieve an EM of around 22\% by predicting none for all test cases, while the F1 can only be near zero. In MATRES, the label class distribution is highly imbalanced, where there are 52.3\% instances with the label `before' and only 2.4\% with the label `unknown'. By only predicting `before' we can reach a micro-F1 of 0.523 while a near-zero score on macro F1, which motivates us to use macro F1 as the primary metric in \Cref{tab:matres_main_results}. In the case when LLMs mostly predict \textit{None} or `before', the most dominant labels in the dataset that will lead to a high EM or micro F1, it makes more sense to use binary F1 or macro F1 as the primary metric.

\section{Additional Details of Knowledge Conflict Diagnosis} \label{sec:appendix_knowledge_conflict_detection}
In TORQUE, for example, 
for the question ``\textit{What happened after Bush gave four key speeches?}'' and answers ``\{\textit{called}, \textit{elect}, \textit{vote}\}'' under a certain context,
we can acquire three event relation triples (\textit{gave}, \textit{before}, \textit{called}), (\textit{gave}, \textit{before}, \textit{elect}), and (\textit{gave}, \textit{before}, \textit{vote}).
and use those triples for calculating and detecting bias.

For each type of bias, we empirically set thresholds to select knowledge-conflict subsets as in \Cref{sec:appendix_bias_param}. 
For instance, in a feature-relation pair $f$ (e.g., $f$ represents dependency) and $r$, it is knowledge-conflict if $b(f, r) < \frac{c(r)}{\sum_{r'\in R}c(r')}$, indicating that it does not conform to the dominant distribution of relation $r$.
Such selection criteria can be further enhanced by setting a threshold $\mathcal{T}_r<\frac{c(r)}{\sum_{r'\in R}c(r')}$, which increases the level of conflicts by further restricting $b(f, r)$ to be less than $\mathcal{T}_r$.

\section{Impact of Knowledge Conflict}

We compare the model performance on knowledge conflict subsets and the non-conflict subsets to show the impact of knowledge conflicts on model performance in \Cref{tab:effect_conflict}.
In general, models perform more poorly on the conflict subsets, compared with those without conflicts. 
This discovery suggests that the chosen conflict subsets pose greater challenges for PLMs and LLMs, thus validating the efficacy of our conflict detection framework.

\section{Additional Details of the Models}\label{sec:appendix_model}

\subsection{Baselines}\label{sec:appendix_baseline}

For TORQUE, the model consists of a one-layered perceptron built on top of RoBERTa. The transformers' output corresponding to the token being analyzed serves as input to the perceptron layer as a sequence tagging task, where the expected output is either 0 or 1, indicating whether this event argument is a correct answer or not. 
Following the original paper of TORQUE, we fine-tuned RoBERTa-large on the training set of TORQUE, using a batch size of 6 (each input is a concatenation of one passage and one question, and the output is a vector measuring the probability of each event argument token).
The learning rate is 1e-5, total epoch is 10, and three random seeds were selected. 
The experiments are conducted on NVIDIA A5000 GPUs, which takes around 30 minutes for training one epoch.

\begin{table}[t]
\centering
\small
\renewcommand\arraystretch{1.1}
\newcolumntype{C}[1]{>{\centering\arraybackslash}p{#1}}
\begin{tabular}{l|cccc}
\toprule
 & \multicolumn{2}{c}{Conflict} & \multicolumn{2}{c}{Non-Conflict} \\
 \cline{2-5}
 & EM & F1 & EM & F1 \\
\midrule
\multicolumn{2}{l}{RoBERTa-large} \\ 
\hline
Rel.Prior & 29.5$\downarrow$ & 73.3$\downarrow$ & 40.7 & 74.5 \\
Rel.Prior (warm-up) & 50.0$\downarrow$ & 75.1$\downarrow$ & 75.0 & 76.2 \\
Narrative & 31.4$\uparrow$ & 69.0$\downarrow$ & 48.4 & 75.2 \\
Tense & 33.5$\downarrow$ & 72.9$\downarrow$ & 50.7 & 75.0 \\
Tense (warm-up) & 48.4$\downarrow$ & 72.4$\downarrow$ & 77.3 & 78.6 \\
Dependency & 41.7$\uparrow$ & 78.6$\downarrow$ & 37.5 & 81.2\\
\midrule
\multicolumn{2}{l}{GPT-3.5} \\ 
\hline
Rel.Prior & 4.82$\downarrow$ & 59.9$\uparrow$ & 4.87 & 51.1\\
Rel.Prior (warm-up) & 4.62$\downarrow$ & 47.0$\uparrow$ & 25.0 & 30.4\\
Narrative & 2.13$\downarrow$ & 50.7$\uparrow$ & 7.21 & 44.4 \\
Tense & 4.46$\downarrow$ & 53.5$\uparrow$ & 7.27 & 42.6 \\
Tense (warm-up) & 5.71$\downarrow$ & 45.9$\uparrow$ & 25.3 & 30.0 \\
Dependency & 2.94$\uparrow$ & 57.7$\uparrow$ & 2.72 & 56.7 \\

\bottomrule
\end{tabular}
\vspace{-0.1in}
\caption{Experimental results on the model performance on knowledge conflict and non-conflict data in TORQUE. The RoBERTa-Large model suffers from performance drop when tested on the conflict subsets. On the contrary, GPT-3.5, when not fine-tuned on the biased training set, suffer less from the knowledge conflict in general. However, there is still a large performance gap on warm-up questions for GPT-3.5, dropping from an EM of around 25\% to 5\%.}
\label{tab:effect_conflict}
\end{table}

In MATRES, each data entry is composed of a passage and the corresponding positions of the two event triggers. 
The model consists of a one-layer perceptron to aggregate the embeddings of the two event triggers provided by the transformers.
We use pre-trained Big Bird~\cite{DBLP:conf/nips/ZaheerGDAAOPRWY20}, a RoBERTa variation that deals with longer documents, following \citet{DBLP:conf/eacl/WangZDGRC23}.
The experiments are conducted on NVIDIA A5000 GPUs, which takes around 2 minutes for training one epoch.

We then introduce the bias-agnostic baselines that we adopt.

\paragraph{PoE~\cite{DBLP:journals/neco/Hinton02} and Learned-mixin~\cite{DBLP:conf/emnlp/ClarkYZ19}.} 
In this line of approaches, a biased model is trained to specifically target biased features in the data. The output of the biased model is then combined with the output of the robust model using product of predicted probabilities. This enables the robust model to focus less on the biased features and improve its overall performance.
Denote the probabilities predicted by the biased model for element $i$ as $b_i$, and the probabilities by the robust model as $p_i$, the ensemble to predict the final label by PoE is:

$$\hat{p_i} = softmax(\log (p_i) + \log (b_i))$$

As PoE assumes conditional independence between the bias in the data and all the features except for bias in the data, which may be too strong, learned-mixin is thus proposed to make the relations between $p_i$  and $b_i$ learnable. 
A function $g(x)$ of the input $x$ is learned to dynamically adjust how much to trust the biased model, leading to the final estimation as:

$$\hat{p_i} = softmax(\log (p_i) + g(x_i) \log (b_i))$$

However, a model could learn to set $ g(x_i) $ to 0 to ignore the effect of biased model, learned-mixin + H is thus proposed by adding an entropy penalty:

$$R = w H(softmax(g(x_i) \log (b_i))$$

Here the entropy function takes the form $H(z) = -\sum_j z_j \log (z_j)$. The entropy term can help encourage the biased term to be non-uniform, providing more biased information.

To train the biased model for all these three baselines, we mask all context except for the event triggers. Other hyperparameters are the same as training a RoBERTa baseline.

\begin{table}[t]
\centering
\small
\renewcommand\arraystretch{1.1}
\newcolumntype{C}[1]{>{\centering\arraybackslash}p{#1}}
\begin{tabular}{l|cc}
\toprule
  & \textbf{TORQUE} & \textbf{MATRES} \\
 \midrule
 Whole Dev Set & 1,483 & 1,000\\
 Rel. Prior (relation)& 88 & 148 \\
 Rel. Prior (warm-up) & 66 & - \\
 Narrative & 140 & 477 \\
 Tense (relation) & 243 & 210 \\
 Tense (warm-up) & 71 & -\\
 Dependency & 68 & 42 \\
\bottomrule
\end{tabular}
\caption{Statistics of each knowledge-conflict subset in TORQUE and MATRES. }
\label{tab:bias_statistics}
\end{table}

\paragraph{Counterfactual Inference~\cite{DBLP:conf/naacl/WangCZCLLYLH22, DBLP:conf/eacl/WangZDGRC23}.}
Counterfactual inference focus on event trigger bias and frequent label bias that leads to spurious correlations. 
A causal graph is established to analyze the causal relations between the effect of event triggers, the whole context, and the final prediction.
To mitigate event trigger bias and label bias, element-wise subtraction operation is conducted to get the final prediction:

$$y = y_x - \lambda_1 y_{\bar{x}, e} - \lambda_2 y_{\bar{x}}$$

where $y_x$ is the prediction given by the model trained on the original data without any masking, $y_{\bar{x}, e}$ is the prediction of the model trained on the data where context except for event triggers are masked, and $y_{\bar{x}}$ is the prediction where the model sees nothing as input, which reflects label bias.
$\lambda_1$ and $\lambda_2$ are tuned by conducting 5-fold cross-validations on the training set. The parameters that yield the best cross validation are selected.
The search space is $[-1, 1]$ with an interval of 0.1.
For TORQUE, $\lambda_1=-0.8, \lambda_2=-0.1$. 
For MATRES, $\lambda_1=-0.1, \lambda_2=0.3$.

\paragraph{AFLite~\cite{sakaguchi2021winogrande, le2020adversarial}.}
AFLITE, which stands for Lightweight Adversarial Filtering, is an alternative bottom-up approach to algorithmic bias reduction proposed by~\cite{sakaguchi2021winogrande}. AFLITE trains an ensemble of linear classifiers on random subsets of the training data and filters other instances in the training data that linear classifiers can correctly classify. The rationale of this baseline is that instances that can be classified correctly by a shallow linear model wound contain artifacts. 

In this paper, we use logistic regression as the linear classifier. We repeat training the logistic regression model 20 times on randomly sampled subsets of the training data. Then, we used the trained logistic regression model to predict the labels of the rest of the training instances. We compute a score for every instance $e$ based on the following equation: 
$$score(e) = \frac{\textit{the times of } e \textit{ is predicted correctly}} {\textit{the times of } e \textit{ is predicted}}.$$
After repeating, we filter instances that owns a score higher than 0.8. Following previouse work~\cite{sakaguchi2021winogrande}, we use dense representations produced by frozen \texttt{robert-large} and \texttt{bigbird-roberta-large}, instead of manually identified lexical features, to train logistic regression classifiers on TORQUE and MATRES, respectively.

\subsection{Large Language Models}\label{sec:appendix_llm}

\paragraph{Prompts for the Tasks.}
For TORQUE, the prompt template we use is ``Q: \{question\}, select none or several from \{all\_events\} $\backslash$n \{context\} $\backslash$n A:''. Here, question, context are provided in each data entry in TORQUE. all\_events indicates all the annotated event triggers in the context. 
GPT3 is expected to generate none or several events that are the answers to the question given the context.
We also check another prompt as an additional analysis, which is ``Given the context \{context\}, \{question\}, select none or several from {all\_events\} $\backslash$n A:''.
The performance analysis are introduced in \Cref{tab:torque_prompt_analysis}.

For MATRES, we formulate the problem as a multi-choice question answering (MCQA) task format, as it's inherently a four-way classification task.
The prompt takes the form ``Given the context:$\backslash$n \{context\} $\backslash$n$\backslash$n Q: What's the temporal relation between the event \{e1\} and \{e2\}? $\backslash$n Choice A: \{e1\} happens before \{e2\}. $\backslash$n Choice B: \{e1\} happens after \{e2\}. $\backslash$n Choice C: \{e1\} happens during \{e2\}. $\backslash$n Choice D: unknown. $\backslash$n Answer only with A, B, C, or D. $\backslash$n$\backslash$n A: Choice''.
Here, e1 and e2 are the target event triggers to be studied. The expected output is either A, B, C, or D.
In addition, we compare our MCQA template with other templates that have been used in previous works, denoted as template 2~\cite{DBLP:journals/corr/abs-2304-14827} and template 3~\cite{DBLP:journals/corr/abs-2304-05454}.
A comparison of different templates are presented in \Cref{tab:matres_prompt_template}.
We also present the effect of the three prompt templates in \Cref{tab:matres_prompt_analysis}, and find that our MCQA template achieves the best performance.

\paragraph{Baselines}

We use In-Context Learning (ICL) and Generative Data Augmentation (GDA) as two intuitive baseline that can be directly comparable to our CDA method.
For ICL, specifically, we retrieve one passage-question pair in TORQUE, and retrieve one example per relation from \textit{before}, \textit{after}, \textit{equal}, and \textit{unknown} as as set of exemplars for MATRES (denoted as 1-shot), to form the ICL demonstration.
Note that ICL is considered few-shot learning while our method is purely zero-shot.
We study the variability of different sets of exemplars as well as the effect of 1-shot and 3-shot ICL in \Cref{tab:matres_prompt_analysis}.
We can find that the performance of ICL is quite stable across different sets of random exemplars, and 3-shot exemplars help on template 1 but not the other two templates.

In addition,  we add an additional baseline named \textit{Generative Data Augmentation} (GDA) that performs exemplar generation without a counterfactual guidance. 
That is to say, we ask LLMs to generate exemplars under all relations from $R$, instead of only under the counterfactual relations. 

\paragraph{Counterfactual Data Augmentation}

We introduce how to do Counterfactual Data Augmentation (CDA) for both PLMs and LLMs.

In CDA for PLM, we generate augmented data at scale. 
For TORQUE, we first retrieve all event pairs that are identified as biased in the training set. 
For an event-relation triple $(e_1, e_2, r)$, where $r$ is identified as knowledge-conflict, which appears less frequently in the training set, we ask Flan-T5 to generate some context where $e_1, e_2$ happens under relation $r$, to augment the undervalued distribution of these two events under the conflict relation $r$.
The prompt is:
``Write a story where $e_1$ happens $r'$ $e_2$:''.
We set temperature as 1 and use greedy decoding to get the results.
After generating the context, 
the question associated with the context is thus $Q$=\textit{What happened $r'$ $e_2$} and the corresponding answer is $e_1$.
We do similar generations for warm-up questions that asks what events have happened / is happening / will happen.
We first acquire events that are knowledge-conflict with regard to a relation $r\in \{\textit{happened, will happen, happening}\}$, and randomly sample two or events that are conflict with regard to $r$. 
We ask Flan-T5 ``Write a story where $e_1$ and $e_2$ $r$''.
The corresponding question associate with the generated context is then $Q$=\textit{What have happened/will happen in the future/is happening?}, based on what $r$ is.
After such augmentations, we conduct an additional filtering step by selecting only knowledge-conflict augmented data.
We keep a proportion of augmented data that is scored with low loss by a fine-tuned PLM on TORQUE to boost the initial learning process when trained on augmented data.
For MATRES, the prompt given to Flan-T5 is ``Write a story where $e_1$ happens $r'$ $e_2$''. Then $r$ is used as the final label.

In CDA for LLM, we generate demonstrations to perform in-context learning.
In MATRES, for an example $(c, e_1, e_2, r)$, we first ask the LLM to predict the temporal relation $r_{LLM}$.
Then we use the same prompt as in CDA for PLM to generate counterfactual examples dedicated to the event pair $(e_1, e_2)$, under relations other than $r_{LLM}$.
The generated examples are thus served as exemplars.
In TORQUE, the pipeline is more complicated. 
An entry is composed of context $c$, the set of event triggers $E$ in $c$, the question $q$, and the answers $a$, which is a subset of $E$.
We first ask an LLM to predict the answers $a_{LLM}$, which is also expected to be a subset of $E$. 
We then ask the LLM itself to generate some context where the ground answers are sampled from $E - a_{LLM}$, using the same prompt as in CDA for PLM.
Examples on MATRES are presented in \Cref{tab:matres_cda_example}.

\section{Additional Analysis and Discussions}\label{sec:appendix_analysis}

\paragraph{Results Analysis on Secondary Metrics} Specifically, in \Cref{tab:torque_main_results} on TORQUE, EM and F1 positively correlate for PLMs intuitively as both scores are fairly high. For LLMs, both ChatGPT and GPT-3.5 perform poorly on the task and tend to predict None (no answers) in a zero-shot setting. This explains the relatively higher EM scores in a zero-shot setting. Incorporating in-context learning leads to both LLMs generating more meaningful predictions, resulting in a corresponding increase in the F1 score. However, the improvement is not substantial enough to surpass the zero-shot setting in terms of EM scores. This is because it is relatively easy to achieve a higher EM score by solely predicting ``None'' for all instances.

In \Cref{tab:matres_main_results} on MATRES, regarding the performance of PLM and LLM, micro and macro and F1 correlates positively for PLM, intuitively, as both micro and macro F1 scores are decently high. On LLMs, GPT-3.5 (\texttt{text-davinci-003}) especially, EM and F1 doesn't positively correlates with each other because GPT-3.5 tends to predict ``before'' for most of the instances (97\%) in a zero shot setting, which contributes to a high Micro-F1 but a low Macro-F1. 
We also studied other prompt templates (as in \Cref{tab:matres_prompt_analysis} in the appendix) and got similar results. This indicates a fairly high label bias of GPT-3.5 on the label `before'. With the involvement of in-context learning, when the labels predicted by GPT-3.5 get more evenly distributed, the macro-F1 significantly improves. However, as the micro-F1 achieved by predicting `before' for all instances is quite high, the micro-F1 cannot be improved to beat the 53.3\% baseline as the improvement is not large enough. In ChatGPT, the predicted labels are not that illy distributed, which makes micro and macro F1 basically positively correlate with each other.

In all, the discrepancy of the EM/F1 trend for PLMs and LLMs is due to the relatively poorer performance of LLMs and the tendency of predicting either ``None'' of ``before'', a form of label bias.

\begin{table}[t]
\centering
\small
\renewcommand\arraystretch{1.1}
\newcolumntype{C}[1]{>{\centering\arraybackslash}p{#1}}
\begin{tabular}{@{}l|llll@{}}
\toprule

Model	& all-EM & all-F1 & \scriptsize{confl-EM} & \scriptsize{confl-F1}\\
\midrule
RoBERTa-L & 50.4 & 75.7 & 39.1 & 73.6 \\
CDA + RoBERTa & 51.0*&76.1&40.5*&74.2\\
GPT3+GDA         &4.85&44.0&3.00&54.5\\
GPT3+CDA         &5.16&44.6&4.22&54.9\\
ChatGPT             &17.7&40.7&6.99&40.0\\
ChatGPT+GDA   &4.38&44.2&2.33&52.6\\
ChatGPT+CDA   &6.72*&45.2*&2.60&53.3*\\

\bottomrule
\end{tabular}
\vspace{-0.1in}
\caption{Significance test on TORQUE. }
\label{tab:significance_matres}
\end{table}

\begin{table}[t]
\centering
\small
\renewcommand\arraystretch{1.1}
\newcolumntype{C}[1]{>{\centering\arraybackslash}p{#1}}
\begin{tabular}{@{}l|llll@{}}
\toprule
Model&\tabincell{c}{micro\\(all)}&\tabincell{c}{macro\\(all)}&\tabincell{c}{micro\\(conf)}&\tabincell{c}{macro\\(conf)}\\
\midrule
RoBERTa-L&70.8&44.9&58.0&31.8\\
CDA+RoBERTa&72.2*&45.5*&59.9*&32.9*\\
GPT-3+GDA&45.6&27.6&34.5&22.6\\
GPT-3+CDA&51.3*&30.0*&35.4&27.9*\\
ChatGPT&39.8&25.9&29.3&21.6\\
ChatGPT+GDA&45.7&30.8&34.7&24.1*\\
ChatGPT+CDA&49.3*&32.0*&36.1*&23.8\\

\bottomrule
\end{tabular}
\vspace{-0.1in}
\caption{Significance test on MATRES. }
\label{tab:significance_matres}
\end{table}

\section{Statistical Significance}

To gain a deeper understanding of the significance of the improvement, we incorporate a statistical test. The result shows that most of the improvements are statistically significant, with some exceptions on GPT-3.5.

We perform a randomization test~\cite{cohen1995empirical} on EM and F1 for Table~\ref{tab:torque_main_results} and \ref{tab:matres_main_results}, indicating significant improvements with p < 0.05 by adding a * to the entries where our model outperforms the baseline. Specifically, in TORQUE, CDA + RoBERTa demonstrates significant improvements under the 'all' and 'conflct.avg.' categories for all-EM with p < 0.05. Moreover, there are significant improvements in F1 and EM from zero-shot to CDA when using ChatGPT. However, for GPT-3.5, the addition of synthetic exemplars results in performance deterioration due to its text generation limitations. Further discussions on this matter can be found in Appendix G (Line 1277-1299).

In MATRES, the improvements from RoBERTa-L+CDA to RoBERTa-L is significant on both ``all'' and `conflict.avg'. In terms of LLMs, adding CDA (inherently also in a zero-shot setting) significantly outperform the zero-shot prompt baseline by a large margin except for `confl-avg-EM'. Compared to it's counterpart GDA, CDA outperforms GDA on overall performance.

\section{Additional Ablations}

In this section, we compare our Counterfactual Data Augmentation method with other popular data augmentation methods to show the effectiveness of CDA with regard to knowledge conflict mitigation.
Specifically, we adopt EDA and Synonym replacement as representative text-editing-based data augmentation baselines, and we use a Generative Data Augmentation (GDA) baseline to automatically generate task data using the same backbone language model, Flan-T5-11B, to generate training data without counterfactual constraints.
The only difference between GDA and CDA is that GDA does not use counterfactual constraints, and GDA can serve as an ablation to study the effect of counterfactual constraints.
The results for TORQUE are presented in \Cref{tab:torque_data_aug} and the results for MATRES are presented in \Cref{tab:matres_data_aug}.

We also study the effect of using Flan-T5-generated data as exemplars for ChatGPT as an additional baseline, as in \Cref{tab:matres_flan_t5}. 
The Macro F1 is improved a tad bit than zero-shot and ICL, while sacrificing the Micro F1 performance.
This indicates that Flan-T5-generated data, even though they are useful for supervised fine-tuning in RoBERTa, cannot be directly used for prompting ChatGPT due to a relatively lower quality.

\begin{table}[t]
\centering
\small
\renewcommand\arraystretch{1.1}
\newcolumntype{C}[1]{>{\centering\arraybackslash}p{#1}}
\begin{tabular}{l|cc}
\toprule
 & Micro F1 & Macro F1 \\
 \midrule 
zero-shot & 39.8 & 25.9 \\
ICL & 43.1 & 23.8\\
Flan-T5-11b ICL & 36.8 & 27.1 \\
GDA & 45.7 & 30.8 \\
CDA & \textbf{49.3} & \textbf{32.0} \\
\bottomrule
\end{tabular}
\vspace{-0.1in}
\caption{Experimental results on MATRES using ChatGPT. }
\label{tab:matres_flan_t5}
\end{table}

\begin{figure}[t]
    \centering
    \includegraphics[width=0.8\linewidth]{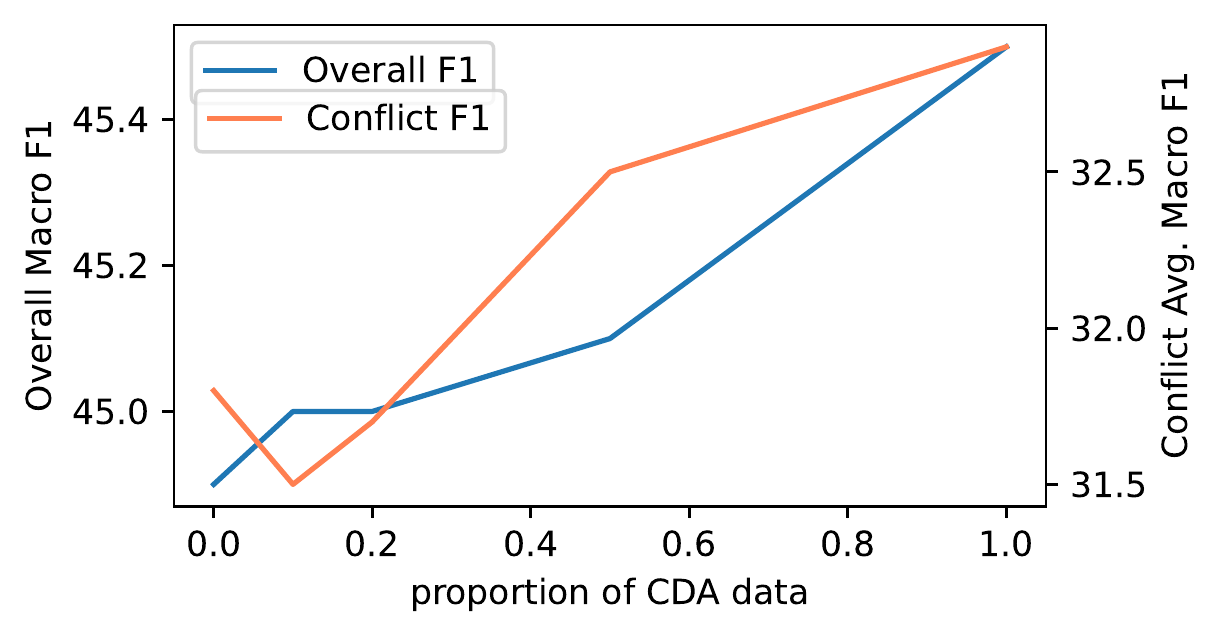}
    \vspace{-0.15in}
    \caption{
    Effect of varying proportions of Counterfactual Data Augmentation (CDA) on MATRES. Models benefit from increased amounts of CDA data.
    }\label{fig:prop_cda}
    \vspace{-1em}
\end{figure}

\begin{table*}[t]
\centering
\small
\renewcommand\arraystretch{1.1}
\newcolumntype{C}[1]{>{\centering\arraybackslash}p{#1}}
\begin{tabular}{p{5.0em}|C{1.2em}C{1.2em}|C{1.4em}C{1.4em}C{1.4em}C{1.4em}C{1.4em}C{1.4em}C{1.4em}C{1.4em}C{1.4em}C{1.4em}C{1.4em}C{1.4em}|C{1.2em}C{1.2em}}
\toprule
 & \multicolumn{2}{c|}{\textbf{all}} & \multicolumn{2}{c}{ \tabincell{c}{\textbf{Rel.Prior}\\ (relation)} } & \multicolumn{2}{c}{\tabincell{c}{\textbf{Rel.Prior}\\(warm-up)}} & \multicolumn{2}{c}{\tabincell{c}{\textbf{Narrative}\\ (relation)}} & \multicolumn{2}{c}{\tabincell{c}{\textbf{Tense}\\(relation)}} & \multicolumn{2}{c}{\tabincell{c}{\textbf{Tense}\\(warm-up)}} & \multicolumn{2}{c|}{\tabincell{c}{\textbf{Dep.}\\(relation)}} & \multicolumn{2}{c}{\tabincell{c}{\textbf{Confl.Avg.}}} \\
 \cline{2-17}
 & EM & F1 & EM & F1 & EM & F1 & EM & F1 & EM & F1 & EM & F1 & EM & F1 & EM & F1  \\
\midrule
RoBERTa-L & \underline{50.4} & 75.7 & {29.5} & 	73.3& 	50.0& 	\underline{75.1}& 	\underline{31.4}& 	\underline{69.0}& 	33.5& 	\underline{72.9}& 	48.4& 	{72.4}& 	41.7& 	78.6 & 39.1 & \underline{73.6}    \\
\ +EDA & 50.2 & 75.5 & \underline{33.5} & \underline{74.2} & \textbf{50.7} & 71.7 & 30.7 & 67.9 & 33.9 & 71.8 & \textbf{50.0} & 69.4 & 41.1 & \textbf{79.6} & \underline{40.0} & 72.4 \\
% \ +Bck.Trans \\
\ +Synonym & 49.7 & \textbf{76.1} & 28.0 & 71.8 & 49.5 & 72.3 & 29.5 & 68.7 & 33.5 & 72.0 & 47.4 & 69.7 & 35.8 & 75.9 & 37.3 & 71.7 \\
% \ +BERT rep. \\
\ +GDA & 49.9 & 75.8 & 30.3 & 73.8 & \underline{50.5} & 74.0 & \textbf{31.7} & \textbf{69.1} & \underline{34.4} & 72.6 & 34.4 & \underline{72.6} & \textbf{49.3} & 71.5 & 38.4 & 72.3 \\
\ +CDA  & \textbf{51.0}&\textbf{76.1}&\textbf{33.7}&\textbf{75.4}&{50.0}&\textbf{75.9}&30.7&68.6&\textbf{35.5}&\textbf{73.1}&\underline{48.8}&\textbf{73.2}&\underline{44.1}&\underline{79.1} & \textbf{40.5} & \textbf{74.2} \\
\bottomrule
\end{tabular}
\vspace{-0.1in}
\caption{Experimental results on the TORQUE dataset using different data augmentation techniques.
Exact-Match (EM) rate and Macro-F1 (F1) scores are reported. 
Best-performed results are \textbf{bold-faced} and the second-best are \underline{underlined}. 
}
\label{tab:torque_data_aug}
\end{table*}

\begin{table*}[t]
\centering
\small
\renewcommand\arraystretch{1.1}
\newcolumntype{C}[1]{>{\centering\arraybackslash}p{#1}}
\begin{tabular}{l|C{2.1em}C{2.1em}|C{2.1em}C{2.1em}C{2.1em}C{2.1em}C{2.1em}C{2.1em}C{2.1em}C{2.1em}|C{2.1em}C{2.1em}}
\toprule
 & \multicolumn{2}{c|}{\textbf{all}} & \multicolumn{2}{c}{ \textbf{Rel. Prior} }  & \multicolumn{2}{c}{\textbf{Narrative}} & \multicolumn{2}{c}{\textbf{Tense}} & \multicolumn{2}{c|}{\textbf{Dependency}} & \multicolumn{2}{c}{\textbf{Confl.Avg.}} \\
 \cline{2-13}
 & Micro & Macro & Micro & Macro & Micro & Macro & Micro & Macro & Micro & Macro & Micro & Macro  \\
\midrule
RoBERTa-large &  {70.8} & 44.9 & 59.7 & 28.5 &   \underline{59.2} & 27.1 & 54.8 & \underline{33.2} &  58.5 & 38.3 & 58.0 & 31.8 \\
\ +EDA & 70.5 & \textbf{46.0} & 60.9 & 29.8 & 58.7 & 27.4 & 55.1 & 33.8 & 60.0 & 38.4 & 58.7 & 32.4 \\
% \ +Bck.Trans \\
\ +Synonym & 70.4 & 45.0 & 59.6 & 28.3 & 59.5 & 26.9 & 55.5 & 33.7 & 61.9 & 41.3 & 57.8 & 32.5 \\
% \ +BERT rep. \\
\ +GDA & \textbf{72.2} & 43.6 &  62.0 & 27.2 & 57.5 & 25.3   & 54.0 & 31.4 & 58.1 & 36.0 & 57.9 & 30.0 \\
\ +CDA (Ours) &  \textbf{72.2} & \underline{45.5} & \textbf{61.5} & {29.3} &  58.8 & {27.3} & {57.2} & \textbf{35.1} &  \textbf{62.2} & \underline{39.9} & \textbf{59.9} & \textbf{32.9} \\
\bottomrule
\end{tabular}
\vspace{-0.1in}
\caption{Experimental results on MATRES using different data augmentation techniques. We use two evaluation metrics, Micro-F1 (denoted as Micro) and  Macro F1 (denoted as Macro). Best-performed results are \textbf{bold-faced} and the second-best are \underline{underlined}.}
\label{tab:matres_data_aug}
\end{table*}

\begin{table}[t]
\centering
\small
\renewcommand\arraystretch{1.1}
\newcolumntype{C}[1]{>{\centering\arraybackslash}p{#1}}
\begin{tabular}{l|cc}
\toprule
 & EM & F1 \\
 \midrule 
CDA (1-shot) & 5.16 & 44.6 \\
CDA (3-shot) & 14.5 & 50.1 \\
 \midrule
template 1 (zero-shot) & 8.36 & 45.5 \\
template 2 (zero-shot) & 8.16 & 45.9 \\
\midrule
template 1 (1-shot)-1 & 4.52 & 43.4 \\
template 1 (1-shot)-2 & 6.00 & 44.7 \\
template 1 (1-shot)-3 & 13.1 & 46.9 \\
template 1 (1-shot)-avg & 7.87 & 45.0 \\
template 2 (1-shot)-1 & 9.51 & 50.5 \\
template 2 (1-shot)-2 & 12.6 & 51.2 \\
template 2 (1-shot)-3 & 10.5 & 48.8 \\
template 2 (1-shot)-avg & 10.9 & 50.2 \\
\hline
template 1 (3-shot)-1 & 13.0 & 46.7\\
template 1 (3-shot)-2 & 16.4 & 48.5 \\
template 1 (3-shot)-3 & 11.2 & 48.2 \\
template 1 (3-shot)-avg & 13.5 & 47.8 \\
template 2 (3-shot)-1 & 19.3 & 56.1 \\
template 2 (3-shot)-2 & 18.6 & 55.4 \\
template 2 (3-shot)-3 & 23.3 & 54.0 \\
template 2 (3-shot)-avg & 20.4 & 55.2 \\
\bottomrule
\end{tabular}
\vspace{-0.1in}
\caption{Experimental results on TORQUE using different prompt templates. }
\label{tab:torque_prompt_analysis}
\end{table}

\begin{table*}
\small
\centering
\begin{tabular}{l|c|c|c|c}
\toprule
\multicolumn{5}{c}{\textbf{MATRES}}                                                   \\
\midrule
\multicolumn{1}{c|}{\textbf{Strategies}} & \textbf{Template input} & \textbf{GPT3.5} & \textbf{Gold} & \textbf{T/F} \\
\midrule
\tabincell{l}{Prompt 1\\ (MCQA)} &
\parbox[c]{8.5cm}{\texttt{Given the context:$\setminus$n} Jim Unruh, Unisys's president, said he is approaching next year with caution. He said the strength of the world-wide economy is suspect, and doesn't see much revenue growth in the cards. He also said that the price wars flaring up in parts of the computer industry will continue through next year. He said the move toward standard operating systems means customers aren't locked into buying from their traditional computer supplier and can force prices down. \texttt{$\setminus$n$\setminus$nQ: What's the temporal relation between the event "}suspect\texttt{" and "}flaring\texttt{"? $\setminus$n Choice A: suspect happens before flaring. $\setminus$n Choice B: suspect happens after flaring. $\setminus$n Choice C: suspect happens during flaring. $\setminus$n Choice D: unknown. $\setminus$Answer only with A, B, C, or D. $\setminus$n$\setminus$nA: Choice} } & A & A & T \\
\midrule
\tabincell{l}{Prompt 2\\\cite{DBLP:journals/corr/abs-2304-14827}}  & \parbox[c]{8.5cm}{\texttt{Determine the temporal order from "}suspect\texttt{" to "}flaring\texttt{" in the following sentence: "}"Jim Unruh, Unisys's president, said he is approaching next year with caution. He said the strength of the world-wide economy is suspect, and doesn't see much revenue growth in the cards. He also said that the price wars flaring up in parts of the computer industry will continue through next year. He said the move toward standard operating systems means customers aren't locked into buying from their traditional computer supplier and can force prices down. "\texttt{". Only answer one word from AFTER, BEFORE, EQUAL, VAGUE. Answer:}}                       & BEFORE                   & BEFORE         & T            \\
\midrule
\tabincell{l}{Prompt 3\\\cite{DBLP:journals/corr/abs-2304-05454}}  & \parbox[c]{8.5cm}{\texttt{Given the document }Jim Unruh, Unisys's president, said he is approaching next year with caution. He said the strength of the world-wide economy is suspect, and doesn't see much revenue growth in the cards. He also said that the price wars flaring up in parts of the computer industry will continue through next year. He said the move toward standard operating systems means customers aren't locked into buying from their traditional computer supplier and can force prices down.\texttt{ and a list of temporal relations [before, after, vague, equal] and event triggers} suspect\texttt{ and }flaring\texttt{. what is the temporal relation between suspect and flaring? Answer vague if unsure. Keep the answer short and concise. }} & before                    &             before & T\\

\bottomrule
\end{tabular}
\caption{Prompt templates for MATRES.}
\label{tab:matres_prompt_template}
\end{table*}

\begin{table*}
\small
\centering
\begin{tabular}{l|c|c|c|c}
\toprule
\multicolumn{5}{c}{\textbf{MATRES}}                                                   \\
\midrule
\multicolumn{1}{c|}{\textbf{Strategies}} & \textbf{Template input} & \textbf{GPT3.5} & \textbf{Gold} & \textbf{T/F} \\
\midrule
\tabincell{l}{Zero-shot} &
\parbox[c]{10cm}{\texttt{Given the context:$\setminus$n} [Context] \texttt{$\setminus$n$\setminus$nQ: What's the temporal relation between the event "}$e_1$\texttt{" and "}$e_2$\texttt{"? $\setminus$n Choice A: $e_1$ happens before $e_2$. $\setminus$n Choice B: $e_1$ happens after $e_2$. $\setminus$n Choice C: $e_1$ happens during $e_2$. $\setminus$n Choice D: unknown. $\setminus$Answer only with A, B, C, or D. $\setminus$n$\setminus$nA: Choice} } & A & B & F \\
\midrule
\tabincell{l}{Counterfactual\\ generation} & \parbox[c]{10cm}{
\texttt{Generate a paragraph where event }{$e_1$}\texttt{ happens before} {$e_2$}:\\
\texttt{Generate a paragraph where event }{$e_1$}\texttt{ happens after}  {$e_2$}:\\
\texttt{Generate a paragraph where event }{$e_1$}\texttt{ happens in the same time as }{$e_2$}:\\
\texttt{Generate a paragraph where the temporal relation of }{$e_1$}\texttt{ and} {$e_2$}\texttt{ cannot be determined based on the context:}} & \tabincell{l}{$c_A$, $c_B$, \\$c_C$, $c_D$} & / & /  \\
\midrule
\tabincell{l}{CDA prompting} & \parbox[c]{10cm}{\texttt{Given the context:$\setminus$n} $c_B$ \texttt{$\setminus$n$\setminus$nQ: What's the temporal relation between the event "} $\cdots$ \texttt{A: Choice B} \\
\texttt{Given the context:$\setminus$n} $c_C$ \texttt{$\setminus$n$\setminus$nQ: What's the temporal relation between the event "} $\cdots$ \texttt{A: Choice C} \\
\texttt{Given the context:$\setminus$n} $c_D$ \texttt{$\setminus$n$\setminus$nQ: What's the temporal relation between the event "} $\cdots$ \texttt{A: Choice D} \\ 
\texttt{Given the context:$\setminus$n} [Context] \texttt{$\setminus$n$\setminus$nQ: What's the temporal relation between the event "}$e_1$\texttt{" and "}$e_2$\texttt{"? $\setminus$n Choice A: $e_1$ happens before $e_2$. $\setminus$n Choice B: $e_1$ happens after $e_2$. $\setminus$n Choice C: $e_1$ happens during $e_2$. $\setminus$n Choice D: unknown. $\setminus$Answer only with A, B, C, or D. $\setminus$n$\setminus$nA: Choice} 
} & B & B & T \\
\bottomrule
\end{tabular}
\caption{A running example of CDA in MATRES. The LLM itself first predict the label of the example, where the prediction is denoted as $r_{LLM}$. Then, the LLM is asked to generate four context given $e_1$ and $e_2$ under four different temporal relations, using the prompts in the second columns, where the corresponding generated context are then $c_A$, $c_B$, $c_C$, $c_D$. 
Then, the generated contexts other than under the predicted relation $r_{LLM}$ are used as demonstrations for in-context learning.
}
\label{tab:matres_cda_example}
\end{table*}

\begin{table}[t]
\centering
\small
\renewcommand\arraystretch{1.1}
\newcolumntype{C}[1]{>{\centering\arraybackslash}p{#1}}
\begin{tabular}{l|cc}
\toprule
 & Micro F1 & Macro F1 \\
 \midrule
 CDA (1-shot) & 51.3 & \textbf{30.0} \\
 CDA (3-shot)$^*$ & 51.5 & 26.3 \\
\midrule
template 1 zero-shot (MCQA) & \underline{53.3} & \underline{19.7}\\
template 2 \cite{DBLP:journals/corr/abs-2304-14827} & 52.1 & 17.1 \\
template 3 \cite{DBLP:journals/corr/abs-2304-05454} & 13.4 & 13.0 \\
\hline
template 1 (1-shot)-1 & 52.3 & 18.5  \\
template 1 (1-shot)-2 & \underline{53.1} & \underline{20.4}  \\
template 1 (1-shot)-3 & 51.6 & 18.4 \\
template 1 (1-shot)-avg & 52.3 & 19.1 \\
template 1 (1-shot)-MV & 52.1 & 19.0\\
template 2 (1-shot)-1 & 49.9 & 22.0  \\
template 2 (1-shot)-2 & 49.3 & 22.1  \\
template 2 (1-shot)-3 & 50.1 & 19.8 \\
template 2 (1-shot)-avg & 49.8 & 21.3 \\
template 2 (1-shot)-MV & 50.0 & 20.6 \\
template 3 (1-shot)-1 &  32.7 & 18.6 \\
template 3 (1-shot)-2 &  34.4 & 20.7 \\
template 3 (1-shot)-3 & 28.8 & 17.8  \\
template 2 (1-shot)-avg & 32.0 & 19.0 \\
template 3 (1-shot)-MV & 31.9 & 18.5 \\
\hline
template 1 (3-shot)-1$^*$ & 57.5 & 24.1  \\
template 1 (3-shot)-2$^*$ & \textbf{57.0} & \underline{28.0} \\
template 1 (3-shot)-3$^*$ & 50.0 & 23.4 \\
template 1 (3-shot)-avg$^*$ & 54.8 & 25.2 \\
template 1 (3-shot)-MV$^*$ & \underline{57.0} & 24.4  \\
template 2 (3-shot)-1$^*$ & 46.5 & 18.2 \\
template 2 (3-shot)-2$^*$ & 47.0 & 18.1\\
template 2 (3-shot)-3$^*$ & 47.5 & 24.9\\
template 2 (3-shot)-avg$^*$& 47.0 & 20.4 \\
template 2 (3-shot)-MV$^*$ & 48.0 & 19.2  \\
template 3 (3-shot)-1$^*$ & 35.5 & 21.3 \\
template 3 (3-shot)-2$^*$ & 29.0 & 15.7 \\
template 3 (3-shot)-3$^*$ & 34.0 & 20.2 \\
template 2 (3-shot)-avg$^*$ & 32.8 & 19.1 \\
template 3 (3-shot)-MV$^*$ & 33.0 & 19.2 \\
\bottomrule
\end{tabular}
\vspace{-0.1in}
\caption{Experimental results on MATRES using different prompt templates. $*$ indicates we test the performance on the same 200 randomly down-sampled examples from MATRES.
We run 3 different random seeds per few-shot in-context learning experiments. `avg' indicates the average between the three runs, and `MV' indicates the majority voting across the three runs.
}
\label{tab:matres_prompt_analysis}
\end{table}

\end{document}